\documentclass{article}
\usepackage{graphicx} 
\usepackage{subfigure} 
\usepackage{algorithm}
\usepackage{algorithmic}
\usepackage{amsmath}
\usepackage{amsthm}
\usepackage{amsfonts}
\usepackage{amssymb}
\usepackage{multirow}

\def\argmax{\mathop{\mathrm{argmax}}}

\newtheorem{lemma}{Lemma}
\newtheorem{definition}{Definition}

\newtheorem{corollary}{Corollary}
\newtheorem{theorem}{Theorem}

\usepackage{times}
\usepackage{helvet}
\usepackage{courier}
\usepackage[numbers]{natbib}

\title{Hybrid Batch Bayesian Optimization}

\author{Javad Azimi\\azimi@eecs.oregonstate.edu\\Oregon State University\\
\and
Ali Jalali\\alij@utexas.edu\\University of Texas at Austin\\
\and
Xiaoli Fern\\xfern@eecs.oregonstate.edu\\Oregon State University
}

\begin{document} 

\maketitle

\begin{abstract}
Bayesian Optimization (BO) aims at optimizing an unknown function that is costly to evaluate. We focus on applications where concurrent function evaluations are possible. In such cases, BO could choose to either sequentially evaluate the function (\emph{sequential} mode) or evaluate the function at a batch of multiple inputs at once (\emph{batch} mode). The sequential mode generally leads to better optimization performance as each function evaluation is selected with more information, whereas the batch mode is more time efficient (smaller number of iterations). Our goal is to combine the strength of both settings. We systematically analyze BO using a Gaussian Process as the posterior estimator and provide a hybrid algorithm that dynamically switches between sequential and batch with \emph{variable} batch sizes. We theoretically justify our algorithm and present experimental results on eight benchmark BO problems. The results show that our method achieves substantial speedup (up to $78\%$) compared to sequential, without suffering any significant performance loss.
\end{abstract}

\section{Introduction}
Bayesian optimization tries to optimize an unknown function $f(\cdot)$ by requesting a set of experiments when $f(\cdot)$ is costly to evaluate \cite{taxonomy, Brochu09}. In this work, we are interested in finding a point $x^* \in \mathcal{X}^d $ such that:
\begin{equation}
x^*=\argmax_{x \in \mathcal{X}^d} f(x),
\end{equation}
where $\mathcal{X}^d$ is our $d$-dimensional compact input space and $f(\cdot)$ is the non-concave underlying function which has multiple local optima. The function $f(\cdot)$ might be the performance of a black box device characterized by input $x$. For example, in our motivating application we try to optimize the power output of nano-enhanced Microbial Fuel Cells (MFCs). MFCs \cite{MFC1} use micro-organisms to generate electricity. It has been shown that efficiency of generated electricity power significantly depends on the surface properties of the anode \cite{MFC3}. Our problem involves optimizing the surface properties of the anodes in order to maximize the output power.  The goal is to develop an efficient BO algorithm for this application since running an experiment is very expensive and time consuming.


Focusing on the task of function maximization, each run of BO consists of two main steps: estimating the values of the unknown function $f(\cdot)$ via a probabilistic model such as GP, and selecting the best next experiment(s) according to the probabilistic model via some selection criterion. The results of the experiment(s) are then be added to update the probabilistic model and this cycle is repeated until we meet a stopping criterion. 

Most of the proposed selection criteria in BO are \emph{sequential}, where only one experiment is selected at each iteration \cite{Moore95,taxonomy,Sacks89,Locatelli97}. Sequential policies usually perform very well in practice, since they optimize the experiment selection at each iteration by using the maximum available information for each experiment. However, they are not time efficient in many applications where running an experiment takes a long time, and we have the capability to run multiple experiments in parallel. This motivates the \emph{batch} algorithms in which more than one experiment is selected at each iteration.

Recently, \citet{azimi10} introduced a \emph{batch} BO approach that selects a batch of $k$ experiments at each iteration that approximates the behavior of a given sequential heuristic. \citet{Gins10} introduced a \emph{constant liar} heuristic algorithm to select a batch of experiments based on the Expected Improvement (EI) \cite{Locatelli97} policy. Specifically, after selecting an experiment by EI, the output of the selected point is set to a constant value. This experiment is then added to the prior and the procedure is repeated until $k$ experiments are selected. Although these two batch algorithms \cite{azimi10,Gins10} can speedup the experiment selection by a factor of $k$, their results show that batch selection in general performs worse than the sequential EI policy, especially when the total number of experiments is small. This observation motivates us to introduce a \emph{Hybrid} BO approach that dynamically alternates between sequential and batch selection to achieve improved time efficiency over sequential without degrading the optimization performance.

In this paper, we focus on a class of batch policies that is based on simulating a sequential policy and provide a systematic approach to analyze such batch BO policies. We analytically connect the mismatch between the BO's probabilistic model and the underlying true function to the performance of the batch policy. We provide full characterization of simulated-based batch policies when the batch size is $2$. 
For the purpose of illustration, consider a batch policy that selects $2$ experiments. The first experiment matches the sequential policy. The choice of the second experiment, however, will depend on what is the simulated outcome of the first experiment. We show that the distance between the second experiment picked by a simulation-based batch policy (without the knowledge of the output of the first experiment) and the one picked by the sequential policy (with the knowledge of the output of the first experiment) is upper-bounded by a quantity that is proportional to the \underline{square root} of the estimation error (of the outcome of the first experiment).

This analysis naturally gives rise to our hybrid batch/sequential algorithm. Our algorithm works as follows: At each step, given any sequential policy (EI in this paper), find the best next single experiment and estimate its possible outcome via BO's probabilistic model (GP in this paper). Then, update the prior with that point and choose the next best single experiment and so on. We analytically show that this process can be continued until a certain stopping criterion is met. This stopping criterion measures how much a simulated experiment is going to bias our probabilistic model (mainly because of inaccuracy in estimation of the outcomes of the first experiment). If the bias is small, we continue to add more examples to our batch; and if it is large, we stop.

The proposed algorithm has the appealing property that it behaves more like a sequential policy in early stages when the number of observed experiments is small, and naturally transits to batch mode in later stages when more experiments are available. This is because the stopping criterion tends to be more stringent in early stages because the bias of the prior can be potentially large, forcing the algorithm to act sequentially.  The beauty of this algorithm is that it evolves from a sequential algorithm to a batch algorithm in an optimal manner characterized by our theoretical results.


Experimental results show that the proposed algorithm can achieve up to $78\%$ speedup over the sequential policy without degrading the performance even with a very small number of experiments. We also show that, by increasing the number of experiments, the speedup rate is increased significantly which is consistent with the theoretical results presented in the paper.

The paper is organized as follows. We introduce the Gaussian Process which is used as our model in Section \ref{sec:gp}. The proposed dynamic batch algorithm is described in Section \ref{sec:dbbo}. Section \ref{sec:exp} presents the experimental results and the paper is concluded in Section \ref{sec:con}

\section{Gaussian Process}
\label{sec:gp} 
A BO algorithm has two main ingredients: a probabilistic model for the unknown function, and, a \emph{selection criterion} for choosing next best experiment(s) based on the model. We select GP \cite{Rasmussen06} as our probabilistic model and EI \cite{Locatelli97} as our selection criterion. We study the properties of GP in this section and postpone the analysis of EI to the next section.

We use GP to build the posterior over the outcome values given our observation set $\mathcal{O}=(\boldsymbol{x}_\mathcal{O},\boldsymbol{y}_\mathcal{O})$, where, $\boldsymbol{x}_\mathcal{O}=\{x_1,x_2,\ldots,x_n\}$ is the set of inputs and $\boldsymbol{y}_\mathcal{O}=\{y_1,y_2,\ldots, y_n\}$ is the set of outcomes (of the experiment) such that $y_j=f(x_j)$ and $f(\cdot)$ is the underlying unknown function.

For a new input point $x_i$, GP models the unknown output $y_i=f(x_i)$ as a normal random variable $y_i\sim\mathcal{N}(\mu_{x_i|\mathcal{O}}, \sigma^2_{x_i|\mathcal{O}})$, with  $\mathbf{\mu}_{x_i|\mathcal{O}}=k(x_i,\boldsymbol{x}_\mathcal{O})k(\boldsymbol{x}_\mathcal{O}, \boldsymbol{x}_\mathcal{O})^{-1}\boldsymbol{y}_\mathcal{O}$ and
$\mathbf{\sigma}_{x_i|\mathcal{O}}^{2}=k(x_i,x_i)-k(x_i,\boldsymbol{x}_\mathcal{O}) k(\boldsymbol{x}_\mathcal{O},\boldsymbol{x}_\mathcal{O})^{-1} k(\boldsymbol{x}_\mathcal{O},x_i)$, where, $k(\cdot,\cdot)$ is any arbitrary \emph{kernel function}.

\begin{definition}
Let $\boldsymbol{x}=\left\{x_1,x_2,\ldots,x_m\right\} \in \mathcal{X}\setminus\boldsymbol{x}_\mathcal{O}$ be any unobserved set of points. Let $\widehat{\boldsymbol{y}}=\{\hat{y}_1,\hat{y}_2,\ldots,\hat{y}_m\}$ be our estimate of their outputs based on GP considering $y_i|\mathcal{O}\sim\mathcal{N}(\mu_{x_i|\mathcal{O}},\sigma^2_{x_i|\mathcal{O}})$. For any new point $z\in\mathcal{X}\setminus\left\{\boldsymbol{x}_\mathcal{O}\cup \boldsymbol{x}\right\}$, let $y_z|\mathcal{O}\sim\mathcal{N}(\mu_{z|\mathcal{O}},\sigma^2_{z|\mathcal{O}})$ and $y_z|\mathcal{O},(\boldsymbol{x},\widehat{\boldsymbol{y}}) \sim\mathcal{N}(\widehat{\mu}_{z|\mathcal{O},\boldsymbol{x}},\widehat{\sigma}^2_{z|\mathcal{O},\boldsymbol{x}})$.
\end{definition}

Under the GP model, the variance of a point $z$ depends only on the location of the observed points and is independent of their outputs, i.e., $\widehat{\sigma}^2_{z|\mathcal{O},\boldsymbol{x}} = \sigma^2_{z|\mathcal{O},\boldsymbol{x}} $. Therefore, we can update the variance of any point $z$ after finalizing our new query set $\boldsymbol{x}$ without the knowledge of their true outputs $\boldsymbol{y} = f(\boldsymbol{x})$. The following theorem characterizes the change in the variance of $z$ if we query $\boldsymbol{x}$.

\begin{theorem}
Assuming $\Delta(\sigma_z):= \sigma_{z|\mathcal{O}}^2-\sigma^2_{z|\mathcal{O},\boldsymbol{x}}$, we have
\begin{equation}
\Delta(\sigma_z)= \left(CA^{-1}\!B^T\!\!-k(z,\boldsymbol{x})\right)D\left(CA^{-1}\!B^T\!\!-k(z,\boldsymbol{x})\right)^T\!\!\!,
\end{equation}
where, \small$B=k(\boldsymbol{x},\boldsymbol{x}_\mathcal{O})$, $A=k(\boldsymbol{x}_\mathcal{O},\boldsymbol{x}_\mathcal{O})$, $C=k(z,\boldsymbol{x}_\mathcal{O})$ and $D=(k(\boldsymbol{x},\boldsymbol{x})-BA^{-1}B^T)^{-1}$.\normalsize
\label{theorem:varchange}
\end{theorem}

From a practical point of view, this theorem enables us to update the variance of $z$ via computing the difference $\Delta(\sigma_z)$ and add it to the previous value. This scheme is much faster than recalculating the variance of $z$ directly. The computational bottleneck of this update is only the matrix inversion in $D$ with complexity $\mathcal{O}(m^3)$, considering the fact that $k(\boldsymbol{x}_\mathcal{O},\boldsymbol{x}_\mathcal{O})^{-1}$ has been computed before, while the complexity of the direct variance computation is $\mathcal{O}\left((n+m)^3\right)$.

The actual expected value $\mu_{z|\mathcal{O},\boldsymbol{x}}$ heavily depends on the true outputs $\boldsymbol{y} = f(\boldsymbol{x})$, which are not available. Without the knowledge of the true outputs, we make an estimation $\widehat{\mu}_{z|\mathcal{O},\boldsymbol{x}}$ based on the GP-suggested output values $\widehat{\boldsymbol{y}}$. We bound this estimation error in the next theorem.\\

\begin{theorem}
Let $\gamma_z=\left\|(k(z,\boldsymbol{x}) - CA^{-1}B^T)D\right\|_2$. Then,

\begin{equation}
\begin{aligned}
\left|\mu_{z|\mathcal{O},\boldsymbol{x}} - \widehat{\mu}_{z|\mathcal{O},\boldsymbol{x}}\right| &\leq \gamma_z\;\;\big\|\boldsymbol{y}-\widehat{\boldsymbol{y}}\big\|_2\\
\left|\mu_{z|\mathcal{O},\boldsymbol{x}} - \mu_{z|\mathcal{O}}\right|\; &\leq \gamma_z\big\|\boldsymbol{y}-\mu_{\boldsymbol{x}|\mathcal{O}}\big\|_2.
\end{aligned}
\nonumber
\end{equation}
\end{theorem}

Here, $\|\cdot\|_2$ is vector 2-norm. This theorem tells us that our estimation error at point $z$ is proportional to the parameter $\gamma_z$, which is known to us \underline{without} the knowledge of $\boldsymbol{y}$. Intuitively, if $\gamma_z$ is small, we would think that our estimation $\widehat{\mu}_{z|\mathcal{O},\boldsymbol{x}}$ is accurate and hence, we can make our decision about the point $z$ without knowing $\boldsymbol{y}$, i.e., before the result of experiment on $\boldsymbol{x}$ returns. This observation tells us that it is possible to do batch BO without a big loss in performance.

{\bf Remark:} If we want to minimize our estimation error of $\widehat{\mu}_{z|\mathcal{O},\boldsymbol{x}}$ in expectation, we should set $\widehat{\boldsymbol{y}}=\mu_{\boldsymbol{x}|\mathcal{O}}$. This is in some sense trivial and even counter intuitive. One might claim that if the unknown function is upper-bounded by $M$, then the best choice for $\widehat{\boldsymbol{y}}$ is $M$ since it increases the expected value around the optimal point in the GP model. However, this theorem shows that this choice is overly optimistic.

The previous theorem provides a performance bound based on our estimation error on $\widehat{\boldsymbol{y}}$, however, from a practical point of view, that bound cannot be computed since we do not know the exact values of $\boldsymbol{y}$. As a practical measure, we would like to focus on the expected value of the estimation error as opposed to the error itself. Next corollary provides an upper-bound on the expected error, by simply taking expectation from the result of theorem 2.\\

\begin{corollary}
Let $\theta_{\boldsymbol{x}} := \sqrt{\sum_{i=1}^m\sigma^2_{x_i|\mathcal{O}}}$, then
\begin{equation}
\mathbb{E}_{\boldsymbol{y}}\big[|\mu_{z|\mathcal{O},\boldsymbol{x}} - \mu_{z|\mathcal{O}}|\big]\leq \gamma_z\theta_{\boldsymbol{x}}.
\nonumber
\end{equation}

Moreover,
\begin{equation}
\mathbb{E}_{\boldsymbol{y}}\left[|\mu_{z|\mathcal{O},\boldsymbol{x}} - \widehat{\mu}_{z|\mathcal{O},\boldsymbol{x}}|\right]\leq \gamma_z\left(\theta_{\boldsymbol{x}}+\|\widehat{y} - \mu_{\boldsymbol{x}|\mathcal{O}}\|_2\right).
\nonumber
\end{equation}
\label{theorem:expchange}
\end{corollary}

{\bf Remark 1:} We focus on the second bound in this corollary, which has two terms. The first term ($\gamma_z\theta_{\boldsymbol{x}}$) measures ``how close" the point $z$ is to $\boldsymbol{x}$. The second term captures the bias of our estimator $\widehat{y}$. According to this corollary, the best choice for $\widehat{y}$ is the mean $\mu_{\boldsymbol{x}|\mathcal{O}}$.

{\bf Remark 2:} This corollary entails that if for some small value of $\epsilon$, we have
\begin{equation}
\gamma_z\left(\theta_{\boldsymbol{x}}+\|\widehat{y} - \mu_{\boldsymbol{x}|\mathcal{O}}\|_2\right)\leq\epsilon,
\label{eq:criterion}
\end{equation}\normalsize
then, we are guaranteed that
\begin{equation}
\mathbb{E}_{\boldsymbol{y}}\left[|\mu_{z|\mathcal{O},\boldsymbol{x}} - \widehat{\mu}_{z|\mathcal{O},\boldsymbol{x}}|\right]\leq\epsilon.
\nonumber
\end{equation}
Since $\gamma_z$ and $\theta_{\boldsymbol{x}}$ are both computable without the knowledge of $\boldsymbol{y}$, this observation motivates us to use this as a stopping criterion for our algorithm to determine if the current estimation bias is too large to continue selecting more examples in the batch. In the nutshell, when we want to query a batch of samples, if this criterion is met, we are sure that our estimation of $\boldsymbol{y}$ is accurate and hence, we do not need to wait for the label of the selected examples before making the next selection.




\section{Hybrid Batch Bayesian Optimization}
\label{sec:dbbo}
In a sequential approach, we query for only one experiment at a time using a selection criterion (policy), mainly because the selection criterion requires the output of the previous query to find the next best one. Suppose we have the capability of running $n_b$ experiments in parallel, and we are limited by the total number of possible experiments $n_l$. At each iteration, the question is whether or not we can query more than one sample to speed up the experimental procedure without losing performance comparing to the sequential approach. 

We use Expected Improvement (EI) as our base sequential selection criterion. Below we provide the formal definition for EI.

\begin{definition}
EI\cite{Locatelli97} at point $x$ with associated GP prediction $y|\mathcal{O}\sim\mathcal{N}(\mu_{x|\mathcal{O}}, \sigma_{x|\mathcal{O}}^2)$ is defined to be
\begin{equation}
\label{eq:mei}
EI(x|\mathcal{O})= \Big(-u\Phi(-u)+\phi(u)\Big)\sigma_{x|\mathcal{O}},
\end{equation}
where, $u=(y_{max}-\mu_{x|\mathcal{O}})/\sigma_{x|\mathcal{O}}$ and $\displaystyle y_{max}=\max_{y_i\in\boldsymbol{y}_\mathcal{O}}\,y_i$\normalsize. Also, $\Phi(\cdot)$ and $\phi(\cdot)$ represent standard Gaussian distribution and density functions respectively.
\end{definition}

Our proposed algorithm selects a batch (possibly one) of samples at each iteration based on the EI policy, where the batch size is dynamically determined at each step. In particular, the algorithm will continue to select more experiments if the condition in \eqref{eq:criterion} is satisfied for the select point $z$.

To explain the algorithm, suppose we are at the beginning of the first round of the algorithm. Thus far, we have observed $\boldsymbol{y}_{\mathcal{O}} = f(\boldsymbol{x}_{\mathcal{O}})$ at some randomly chosen sample points $\boldsymbol{x}_{\mathcal{O}}$. To form our batch query, we start from an empty set of samples and gradually add the next best sample one at a time. The first sample we pick ($x_1$) is identical to the first sample that sequential EI picks ($x_1^*$), simply because both maximize the same objective, i.e., $x_1=x_1^*$. To pick our second sample, we \underline{estimate} $y_1^*=f(x_1^*)$ by some value $\hat{y}_1$. This estimation, changes the $EI$ function of all unobserved points to some $\widehat{EI}$ function formulated as
\begin{equation}
\widehat{EI}(z|\mathcal{O},x_1^*) = \Big(-\widehat{u}\Phi(-\widehat{u})+\phi(\widehat{u})\Big)\sigma_{z|\mathcal{O},x_1^*},
\nonumber
\end{equation}
where, $\widehat{u}=\frac{\max(y_{max},\hat{y}_1)-\widehat{\mu}_{z|\mathcal{O},x_1^*}} {\sigma_{z|\mathcal{O},x_1^*}}$. This is different from the true EI function:
\begin{equation}
EI(z|\mathcal{O},x_1^*) = \Big(-u\Phi(-u)+\phi(u)\Big)\sigma_{z|\mathcal{O},x_1^*},
\nonumber
\end{equation}
where, $u=\frac{\max(y_{max},y_1^*)-\mu_{z|\mathcal{O},x_1^*}} {\sigma_{z|\mathcal{O},x_1^*}}$. Obviously, optimizing $\widehat{EI}$ might not lead to the optimum of the true $EI$. However, the next lemma shows that these two functions are close to each other for a good estimation $\hat{y}_1$.

\begin{lemma}
\label{lem:EIbound}
At any point $z$, we have

\vspace{-0.3cm}
{\small\begin{equation}
\Big|EI(z|\mathcal{O},x_1^*) - \widehat{EI}(z|\mathcal{O},x_1^*)\Big|  \leq \frac{1}{2}\left(1+\frac{\sigma_{z|\mathcal{O}}}{\sigma_{x_1^*|\mathcal{O}}}\right) \Big|\hat{y}_1-y^*_1\Big|.
\end{equation}}
\end{lemma}
\vspace{-0.1in}

In the light of this lemma, there is hope that $x_2 = \arg\max\,\widehat{EI}$ (a potential batch sample from our algorithm) is close to $x_2^*=\arg\max\,EI$ (the optimal sample picked by sequential policy). The next theorem bounds the error of our algorithm in terms of the second selected point in comparison to the sequential EI.

\begin{theorem}
Let $\Sigma_{\min}$ be the minimum singular value of the Hessian matrix $\frac{d^2\widehat{EI}}{dx^2}(x)$ on the line intersecting $x_2$ and $x_2^*$. Then,

\vspace{-0.3cm}
{\small\begin{equation}
\Big\|x_2^* - x_2\Big\|_2^2 \leq \frac{2}{\Sigma_{\min}}\left(1+\frac{\max(\sigma_{x_2|\mathcal{O}},\sigma_{x_2^*|\mathcal{O}})}{\sigma_{x_1^*|\mathcal{O}}}\right) \;\Big|\hat{y}_1-y^*_1\Big|.
\end{equation}}
\end{theorem}
\vspace{-0.1in}

Here $x_2$ is the second point selected by our simulation based batch method without knowing the outcome of $x_1$, whereas $x_2^*$ is the second point selected by the sequential EI method after knowing the outcome of $x_1$.

{\bf Remark 1:} The parameter $\Sigma_{\min}$ captures the curvature of the $\widehat{EI}$ function around its optimal point $x_2$. This curvature cannot be zero unless $x_2^*$ is very far from $x_2$, which is very unlikely due to the closeness of their expected values (see Corollary 1).

{\bf Remark 2:} This theorem shows that the sample estimation error is proportional to the square root of the estimation error of $y_1^*$. This means that the sample estimation is more sensitive to the output estimation error for functions taking value in $[0,1]$.

This line of analysis can be extended to next samples. These results show that an algorithm based on the estimation can be successful. In practice, after we optimized $\widehat{EI}$ for $x_2$, then, we check the condition \eqref{eq:criterion} (i.e., $\gamma_{x_2}(\theta_{x_1^*}+\|\hat{y}_1-\mu_{y|\mathcal{O}}\|_2)\leq \epsilon$) and if this condition is satisfied, we add $x_2$ to our batch query and move on to $x_3$ and so on. Algorithm \ref{alg:SBEI} summarizes our proposed method for hybrid batch Bayesian optimization.

\begin{algorithm}
\caption{Hybrid Batch Expected Improvement }
\label{alg:SBEI}
{\bf Input:} \small Total budget of experiments ($n_l$), maximum batch size ($n_b$), the predictor ($\widehat{y}$), current observation $\mathcal{O}=(\boldsymbol{x}_\mathcal{O}$,$\boldsymbol{y}_\mathcal{O}$) and stopping threshold $\epsilon$.
\small{
\begin{algorithmic}

		\WHILE {$n_l>0$}
			\STATE $\displaystyle x^*_1\gets\arg\;\max_{x\in \mathcal{X}}\;\; \mbox{$EI$}(x|\mathcal{O})$.
			\STATE $\mathcal{A}\gets(x^*_1,\hat{y}_1)$, $\quad n_l\gets n_l-1$.\vspace{0.2cm}
			\STATE $\displaystyle z\gets\arg\;\max_{x\in \mathcal{X}}\;\; \mbox{$\widehat{EI}$}(x|\mathcal{O}\cup\mathcal{A})$.\vspace{0.1cm}
			\WHILE{ $\left(\gamma_z(\theta_{\boldsymbol{x}_\mathcal{A}}+\|\hat{y}_\mathcal{A} - \mu_{\boldsymbol{x}_\mathcal{A}|\mathcal{O}}\|_2)\leq \epsilon\right)$ and $(n_l>0)$ and $(|\mathcal{A}|< n_b)$}\vspace{0.1cm}
				\STATE $\mathcal{A}\gets\mathcal{A}\cup(z,\hat{y}_z)$, $\quad n_l\gets n_l-1$.\vspace{0.1cm}
				\STATE $\displaystyle z\gets\arg\;\max_{x\in \mathcal{X}}\;\; \mbox{$\widehat{EI}$}(x|\mathcal{O}\cup\mathcal{A})$.
			\ENDWHILE \vspace{0.1cm}
			\STATE $\boldsymbol{y}_\mathcal{A}\gets  \text{RunExperiment}(\boldsymbol{x}_\mathcal{A})$
			\STATE $\mathcal{O}\gets \mathcal{O}\cup (\boldsymbol{x}_\mathcal{A},\boldsymbol{y}_\mathcal{A})$
		\ENDWHILE
	  \STATE  \textbf{return}  $\max(\boldsymbol{y}_\mathcal{O})$
\end{algorithmic}}
\end{algorithm}
\vspace{-0.1in}

In early stages, this algorithm behaves more like a sequential policy since the criterion for building up a batch is very hard to satisfy, mainly because $\theta_{\boldsymbol{x}}$ is large when we have only a few samples in $\mathcal{O}$. After collecting enough samples, the term $\theta_{\boldsymbol{x}}$ starts decreasing and as it gets closer and closer to zero, we can select larger and larger batch sizes. Thus, the algorithm gradually transits into a batch policy while maintaining a close match to the performance to the pure sequential policy.

\section{Experimental Results}\label{sec:exp}
\begin{figure}[h]
\begin{center}
\addtolength{\tabcolsep}{-0.7em}
\begin{tabular}{cc}
\includegraphics[width=1.3in, height=1.3in]{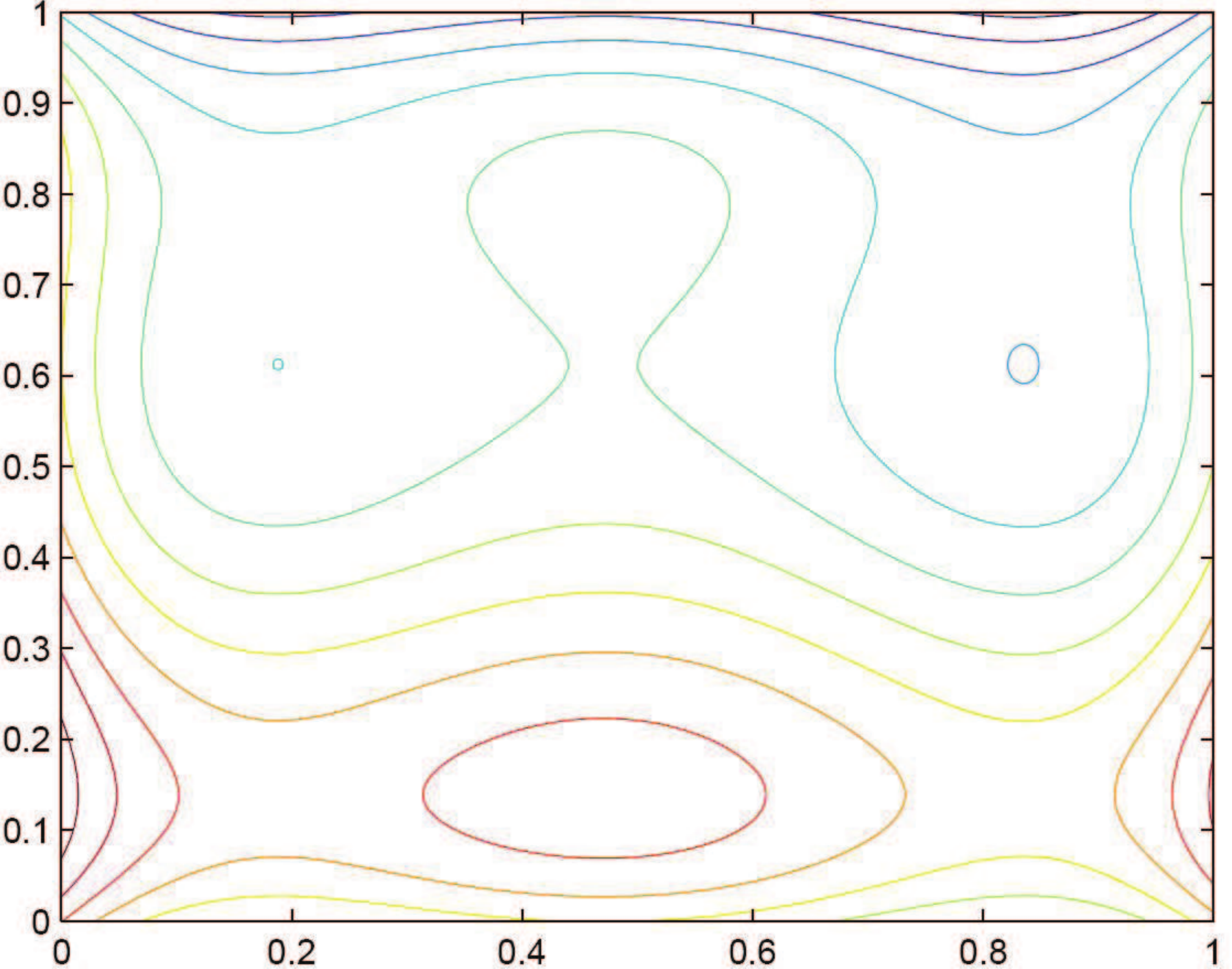}\space\space&
\includegraphics[width=1.3in, height=1.3in]{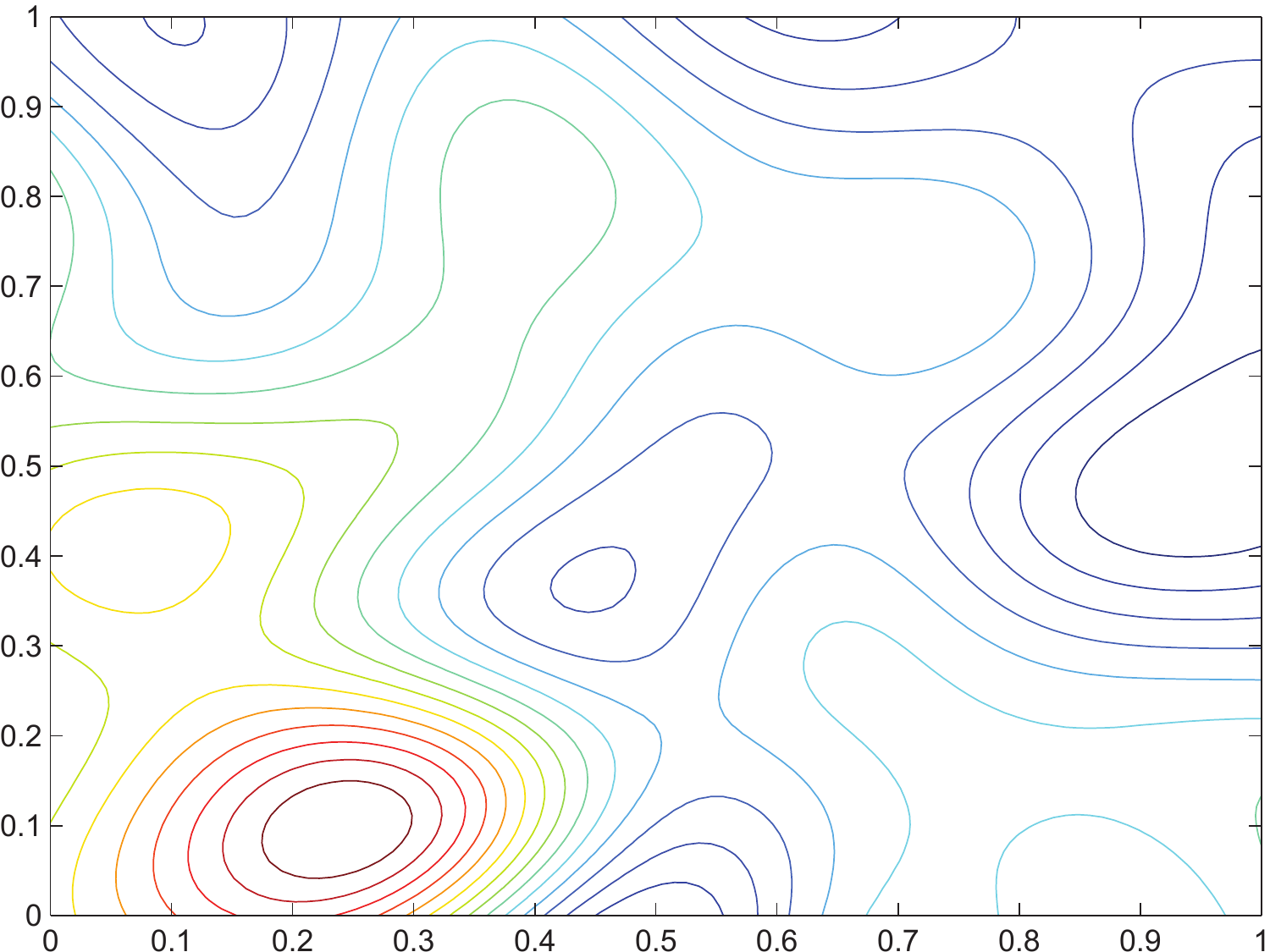}\\
Fuel Cell & Hydrogen\\
\end{tabular}
\end{center}
\caption{The contour plot for FuelCell and Hydrogen.}
\label{fig:contour}
\end{figure}

\textbf{Benchmarks.} We consider  $6$ well-known synthetic benchmark functions: \emph{Cosines} and \emph{Rosenbrock} \cite{Anderson00,Brunato06} over $[0,1]^2$, \emph{Hartman(3)}\cite{GOP78} over  $[0,1]^3$, \emph{Hartman(6)}\cite{GOP78} over  $[0,1]^6$, \emph{Shekel}\cite{GOP78} over  $[3,6]^4$ and \emph{Michalewicz} \cite{Michal94} over $[0,\pi]^5$. The analytic expression for these functions are shown in Table \ref{tab:functions}.

The other two real benchmarks are  \emph{Fuel Cell} and \emph{Hydrogen}. In Fuel Cell, the goal is to maximize the generated electricity from microbial fuel cells with by changing the nano structure properties of the anodes. We fit a regression model on the data to build our function $f(\cdot)$ for evaluation. In Hydrogen benchmark, the data has been collected as part of a study on Hydrogen production from a particular bacteria where the goal is to maximize the amount of Hydrogen production by optimizing the PH and Nitrogen levels of growth medium. Both Fuel cell and Hydrogen data are in $[0,1]^2$. Their contour plots are shown in Figure \ref{fig:contour}.

\begin{table*}[t]
\caption{Benchmark Functions}\label{tab:functions}
\begin{center}
\footnotesize{
\begin{tabular}{|l|c||l|c|}
\hline
\multirow{2}{*}{Cosines(2)} & \footnotesize $1\!-(u^2\!+v^2\!-0.3\cos(3\pi u)\!-0.3\cos(3\pi v))$ & \multirow{2}{*}{Rosenbrock(2)} & \multirow{2}{*}{\footnotesize$10\!-\!100(y\!-x^2)^2\!\!-\!(1\!-x)^2$}\\
&\footnotesize $u=1.6x-0.5, v=1.6y-0.5$ & & \\\hline
\multirow{2}{*}{Hartman(3,6)} & \footnotesize$\sum_{i=1}^4 \Omega_i \exp\left(-\!\sum_{j=1}^d A_{ij}(x_j-P_{ij})^2\right)$ & \multirow{2}{*}{Michalewicz(5)}& \multirow{2}{*}{\footnotesize$-\sum_{i=1}^{5}\sin(x_i)\sin\left(\frac{i\, x_i^2}{\pi}\right)^{\!\!20}$}\\
 & \footnotesize $\Omega_{1\times 4}, \; A_{4 \times d},\; P_{4\times d}$ are constants &  & \\\hline
Shekel(4)& \multicolumn{3}{c|}{ \footnotesize$\sum_{i=1}^{10}\frac{1}{\omega_i+\Sigma_{j=1}4(x_j-B_{ji})^2}$ \qquad $\omega_{1\times 10}, \; B_{4a \times 10}$ are constants}\\ \hline
\end{tabular}}
\end{center}
\end{table*}

\textbf{Setting.} We use a GP using a zero-mean prior and Gaussian kernel function $k(x,y)= \text{exp}( - \frac{1}{l}\parallel x - y \parallel ^2)$, with kernel width $l=0.01\Sigma_{i=1}^d l_i$, where, $l_i$ is the length of the $i^{th}$ dimension \cite{azimi10}. For this kernel function, we can directly drive the next two corollaries from theorems 1, 2.\\

\vspace{-0.1in}
\begin{corollary}
For all points $z\in\mathcal{X}\setminus\left\{\mathcal{O}, x_1^*\right\}$, and kernel function $k(x,y)=e^{-\frac{\parallel x-y\parallel^2}{l}}$, we have $\Delta(\sigma_z)\geq\epsilon$ if

\vspace{-0.1cm}
{\small\begin{equation}
\parallel z-x_1^* \parallel^2\leq -l \ln \Big( \sqrt{n}\parallel A^{-1}B^T\parallel_{2} +\sigma_{x_1^*|\mathcal{O}}\sqrt{\epsilon}\Big).
\nonumber
\end{equation}}
\label{corol:var}
\end{corollary}
\vspace{-0.1in}

This corollary entails that after selecting the first experiment $x_1^*$, the set of points $z$ such that $\Delta(\sigma_z)\geq\epsilon$ are located inside a hyper sphere centered at $x_1^*$. In other words, those inside the hyper sphere are those whose variance is affected significantly (more than $\epsilon$) when $x_1^*$ is selected.\\

\vspace{-0.1in}
\begin{corollary}
Under the assumption of Corollary \ref{corol:var}, we have $\mathbb{E}[|\mu_{z|\mathcal{O},\boldsymbol{x}} - \widehat{\mu}_{z|\mathcal{O},\boldsymbol{x}}|] \geq\epsilon$ if

\vspace{-0.1cm}
{\small\begin{equation}
\parallel z-x^*_1\parallel^2\leq -l\ln \sqrt{\frac{\pi\epsilon^2}{2\sigma^6_{x_1^*|\mathcal{O}}}-n\parallel A^{-1}B^T\parallel_{2}^2}.
\nonumber
\end{equation}}
\label{cor:exp}
\end{corollary}
\vspace{-0.1in}
Similar to corollary \ref{corol:var}, the corollary \ref{cor:exp} represents a hyper sphere centered at $x_1^*$ and the points which are inside the hyper sphere are those whose expected values are affected more than $\epsilon$ when $x_1^*$ is selected.

\begin{table*}[t]
\caption{Benchmarks Performance}\label{table:perf}
\begin{center}
\footnotesize{
\begin{tabular}{l|c|c|c|c|c|c|c|c}
\hline
 & \bf Cosines&\bf Hydrogen&\bf FC&\bf Rosenbrock&\bf Hartman 3&\bf Michalewicz&\bf Shekel& \bf Hartman 6 \\
\hline
\bf Sequential&$0.223$	& $0.048$	&$0.211$	&$0.013$	&$0.042$	&$0.431$	&$0.389$	&$0.263$	\\
\hline
\bf Random&	$0.490$	&$0.282$	&$0.307$	& $0.485$	&	$0.206$&	$0.607$& $0.680$& $0.505$\\
\hline \hline
\bf $\hat{y}=M$&	$0.223$&	$0.048$&$0.211$	&$0.014$	&$0.040$	&$0.429$	&$0.386$	&	$0.270$  \\
\bf Speedup	&$2\%$	&$4\%$	&	$3\%$&$3\%$	&$2\%$	&$2\%$	&	$10\%$ &$2\%$ \\
\hline
\bf $\hat{y}= (1+\zeta)y_{\max}$&$0.222$	&	$0.049$&$0.214$	&$0.012$	&$0.044$	&$0.438$	&$0.401$	&$0.263$	  \\
\bf Speedup	&$22\%$	&$14\%$	&$5\%$	&$10\%$	&$6\%$	&$7\%$	&	$19\%$ &$7\%$ \\
\hline
\bf $\hat{y}=y_{max}$&$0.210$	&$0.050$	&$0.219$	&$0.013$	&$0.040$	&$0.440$	&$0.375$	&$0.276$	  \\
\bf Speedup	&$23\%$	&	$15\%$&$5\%$	&$10\%$	&$11\%$	&$12\%$	&$25\%$	&$13\%$  \\
\hline
\bf $\hat{y}=\hat{\mu}$&	$0.222$&$0.050$	&	$0.214$&$0.011$	&$0.052$	&$0.450$	&$0.412$ &$0.271$		  \\
\bf Speedup	&$45\%$	&$57\%$	&$43\%$	&$37\%$	&$70\%$	&$77\%$	&$78\%$& $75\%$	  \\
\hline
\bf $\hat{y}=y_{min}$&$0.212$	&$0.050$	&$0.213$	&$0.011$	&$0.067$	&$0.444$	&$0.430$	&$0.283$	  \\
\bf Speedup	&$38\%$	&$50\%$	&$32\%$	&$18\%$	&$54\%$	&$75\%$	&$77\%$&$72\%$	  \\
\hline
\bf $\hat{y}=random$&$0.212$	&$0.050$	&$0.211$	&$0.012$	&$0.047$	&$0.440$	&$0.382$ &$0.284$	  \\
\bf Speedup	&$39\%$	&$38\%$	&$20\%$	&$20\%$	&$47\%$	&$58\%$	&$60\%$& $58\%$	  \\
\hline \hline
Matching & 0.295	& 0.085	&0.246	& 0.012	& 0.078	&0.430	&0.521	&0.320\\ \hline
CL($\hat{\mu}$)& 0.301	& 0.084	& 0.257	& 0.012	& 0.081	&0.451	&0.551	&0.319\\ \hline
\end{tabular}}
\end{center}
\end{table*}

We run our algorithm on each benchmark for $100$ independent times and the average \emph{simple regret} is reported as the result. The simple regret is the difference between the maximum value of $f(\cdot)$, denoted by $M$, and $y_{max}$ after finishing the experimental procedure. In each run, the algorithm starts with $2$ initial random points for $2,3$-dimensional benchmarks and $5$ initial random points for higher dimensional benchmarks. The total number of experiments $n_l$ is set to $15$ for $2,3$-dimensional and $30$ for the higher dimensional benchmarks. The maximum batch size at each iteration, $n_b$, is set to $5$. The parameter $\epsilon$ is set to $0.02$ for $2,3$-dimensional and $0.2$ for higher dimensional benchmarks. Note that, our experimental setup is designed to match typical scenarios encountered in real applications, where we typically start with a very small number of random experiments, and are restricted with a total budget.

\textbf{Results.} Our algorithm requires us to select a specific estimation for $\hat{y}$. Recall that our theoretical analysis from Theorem 2 suggests that to minimize the estimation error of $\widehat{\mu}_{z|\mathcal{O},\boldsymbol{x}}$ in expectation, we should use $\widehat{\boldsymbol{y}}=\mu_{\boldsymbol{x}|\mathcal{O}}$. Here we hope to confirm this by comparing different possible estimations for $\hat{y}$. In particular, we consider $6$ different estimations of $\hat{y}$ including:
1) $\hat{y}=M$, which means we expect to observe the best possible output for each experiment selected by EI;
2) $\hat{y}=y_{max}$, where $y_{max}=\max_{y_i\in\boldsymbol{y}_{\mathcal{O}}}\;y_i$ is our current best observation;
3) $\hat{y}=(1+\zeta)y_{max}$, which means each step of EI algorithm is expected to improve the best current observation by margin $\zeta$, we set the value of $\zeta$ to $0.1$ in our experiment;
4) $\hat{y}=\widehat{\mu}_{x|\mathcal{O}}$, which means we set the value of $\hat{y}$ to be the expected output at that point;
5) $\hat{y}=y_{min}$, where $y_{min}=\min_{y_i\in\boldsymbol{y}_{\mathcal{O}}}\;y_i$ is the current minimum observed output; and 6) $\hat{y}=random$, which set $\hat{y}$ to a uniform random value drawn in $[y_{min},y_{max}]$.

To demonstrate the effectiveness of our algorithm, we consider two state-of-the-art batch BO algorithms in the literature: 1) simulation matching (Matching) \cite{azimi10} and 2) the constant liar approach in which the output of the selected samples in the batch is set to their mean in order to select the next experiment (CL($\hat{\mu}$)) \cite{Gins10}. For both methods, we set the batch size to $k=5$. We have also reported the performance of the \emph{sequential EI} and pure random selection policies.

\begin{figure*}
\begin{center}
\begin{tabular}{@{}c@{} @{\ }c@{} @{\ }c@{} @{}c@{}}
\includegraphics[width=1.5 in,height=1.35 in]{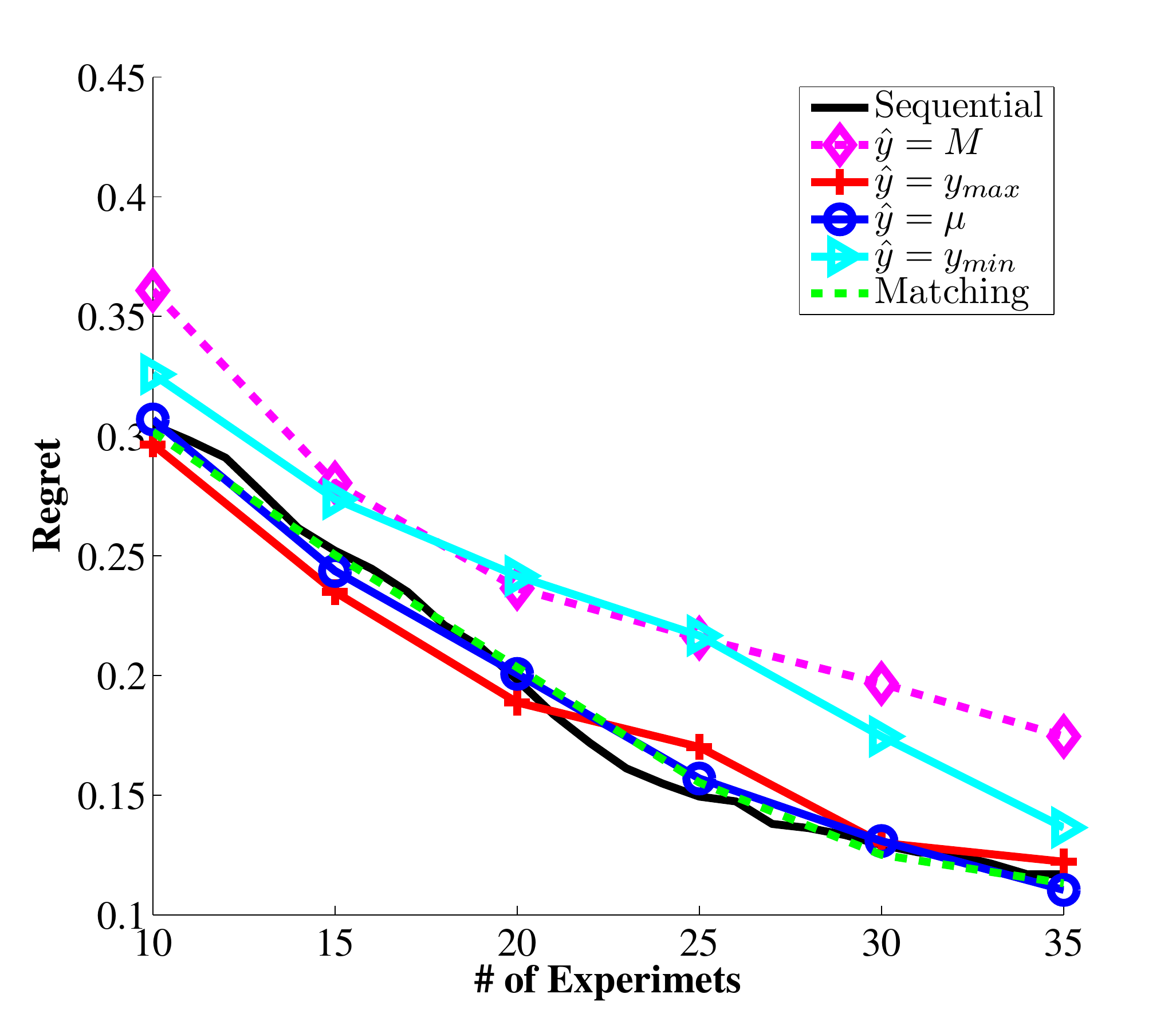} &
\includegraphics[width=1.5 in,height=1.35 in]{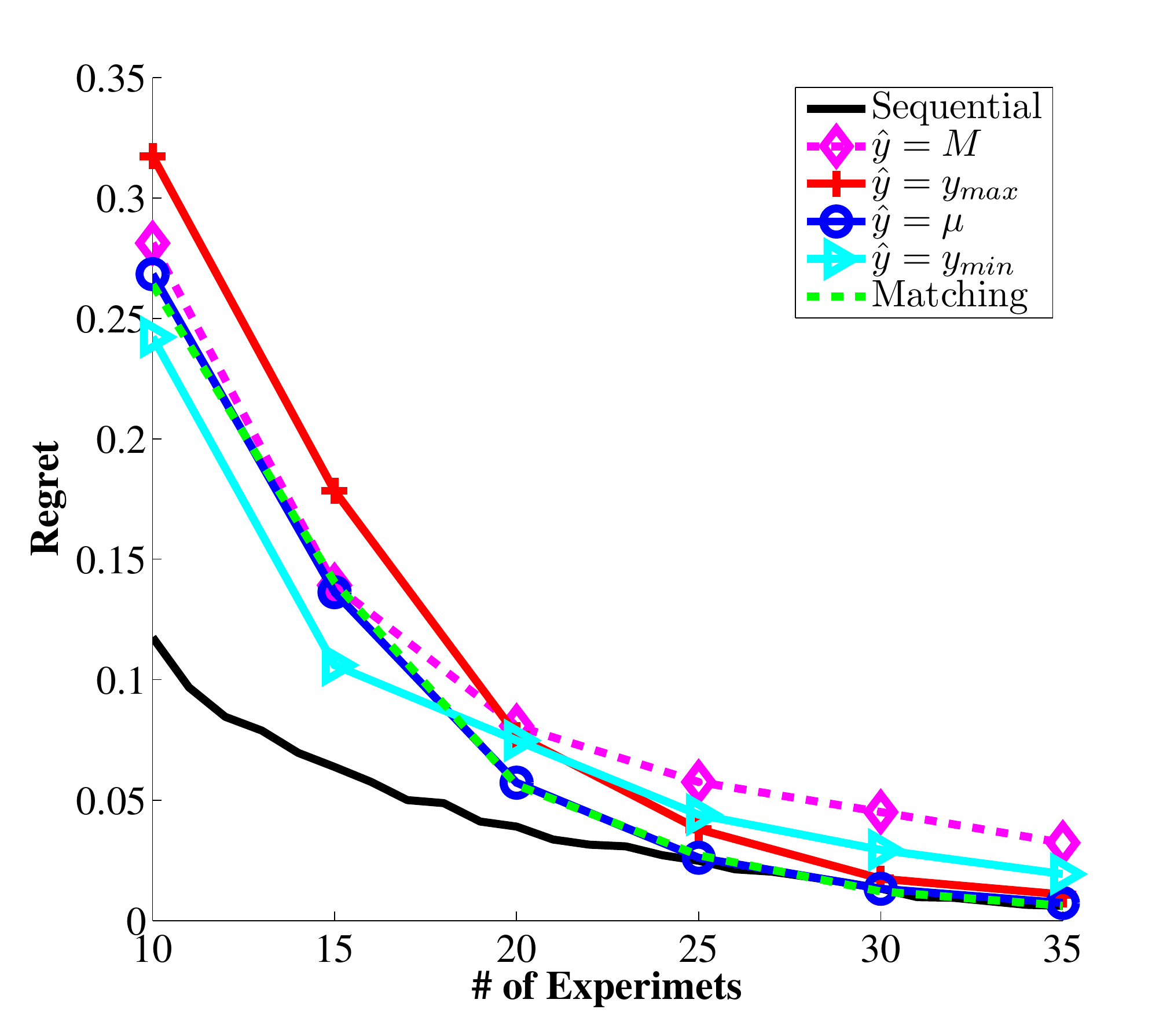} &
\includegraphics[width=1.5 in,height=1.35 in]{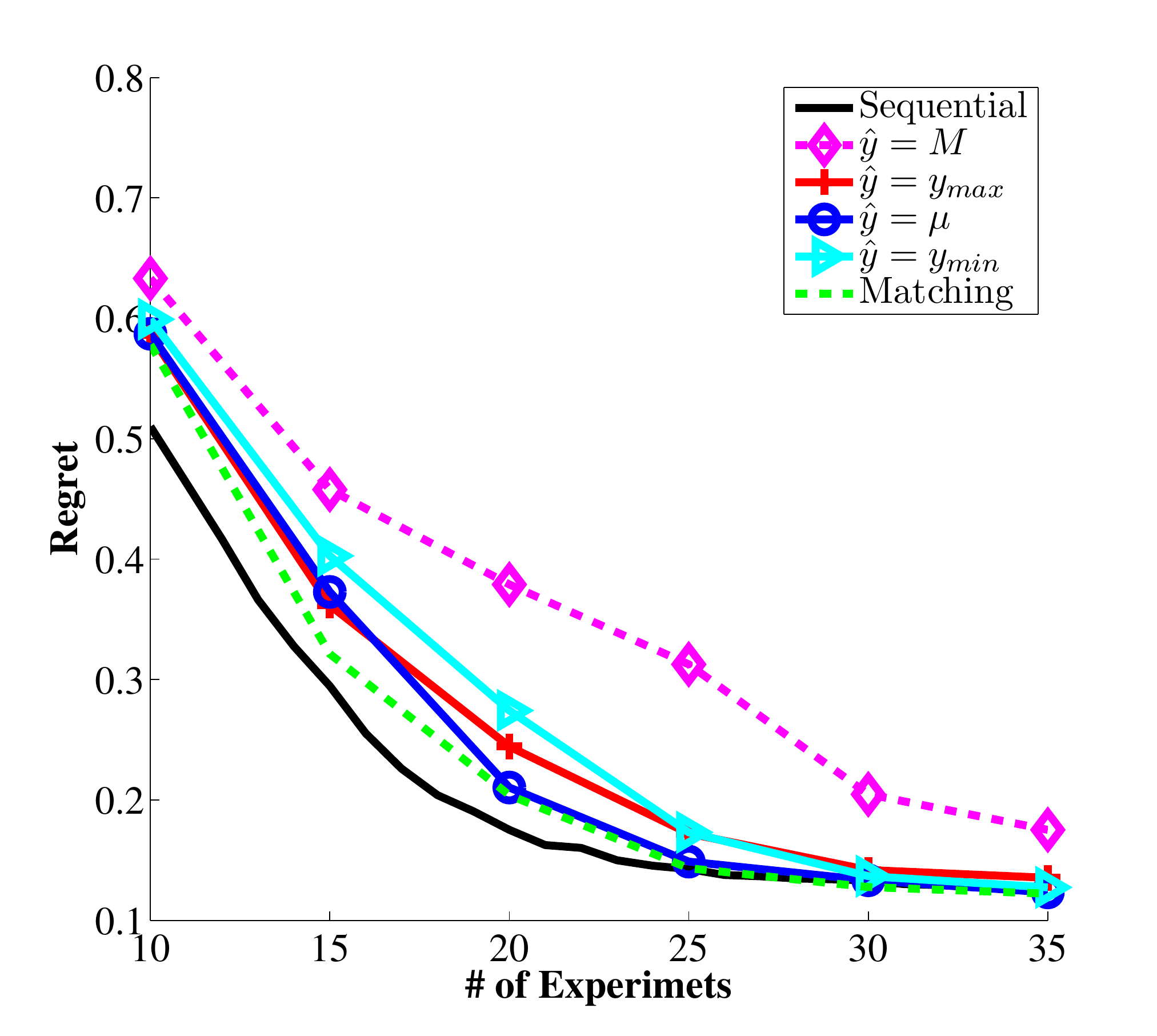} &
\includegraphics[width=1.5 in,height=1.35 in]{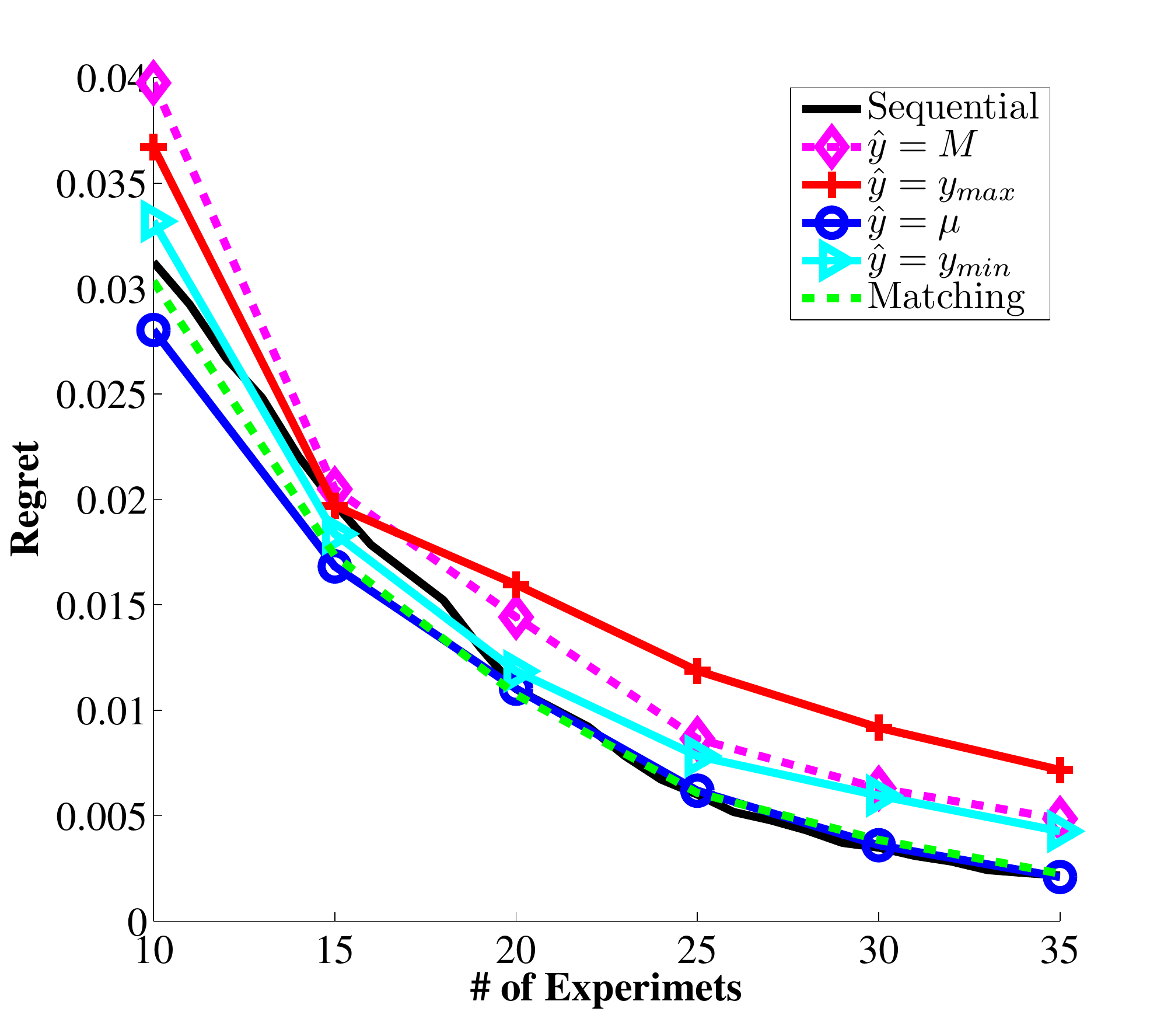} \\
Fuel Cell &Hydrogen &Cosines& Rosenbrock\\
\includegraphics[width=1.5 in,height=1.35 in]{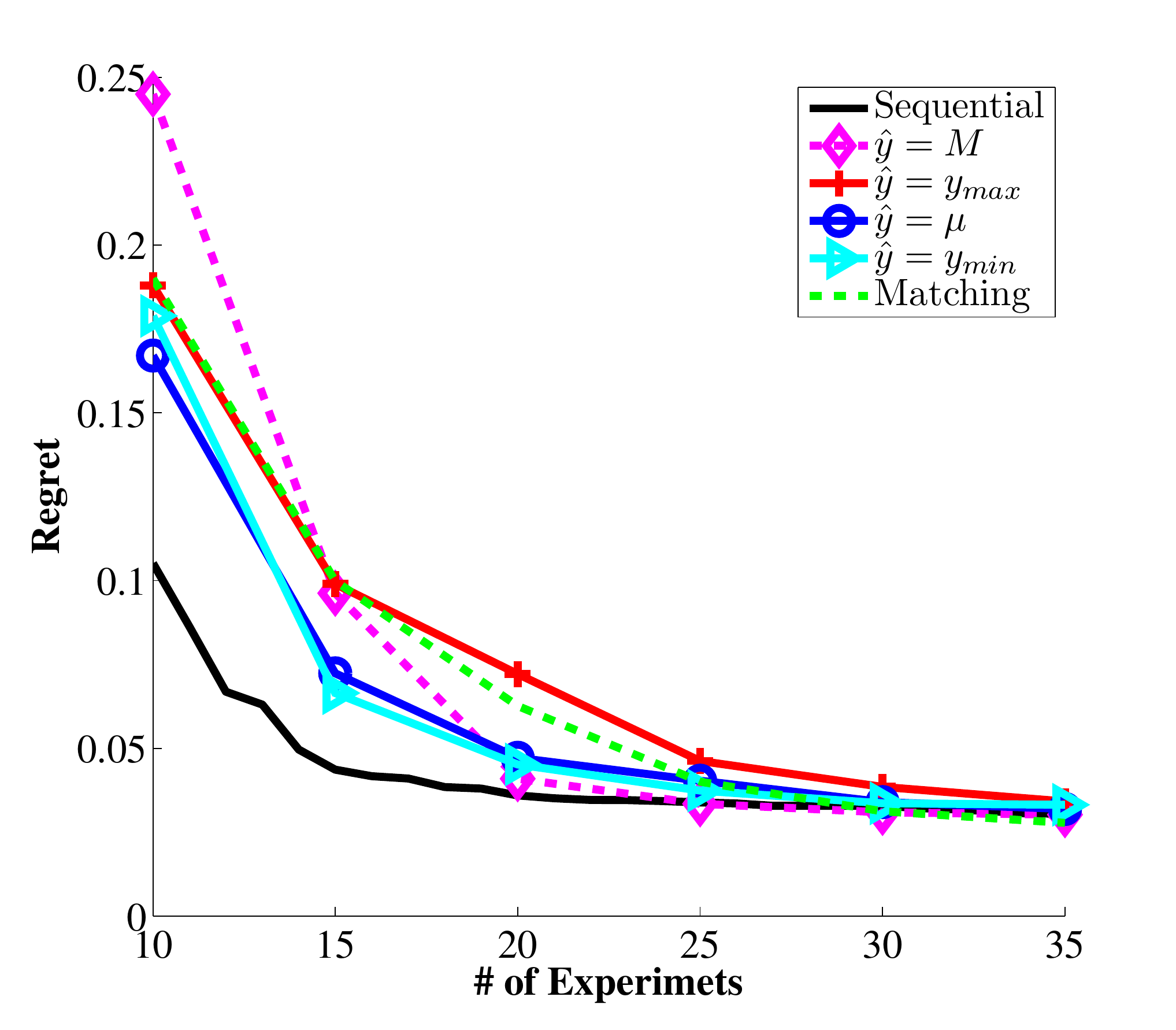} &
\includegraphics[width=1.5 in,height=1.35 in]{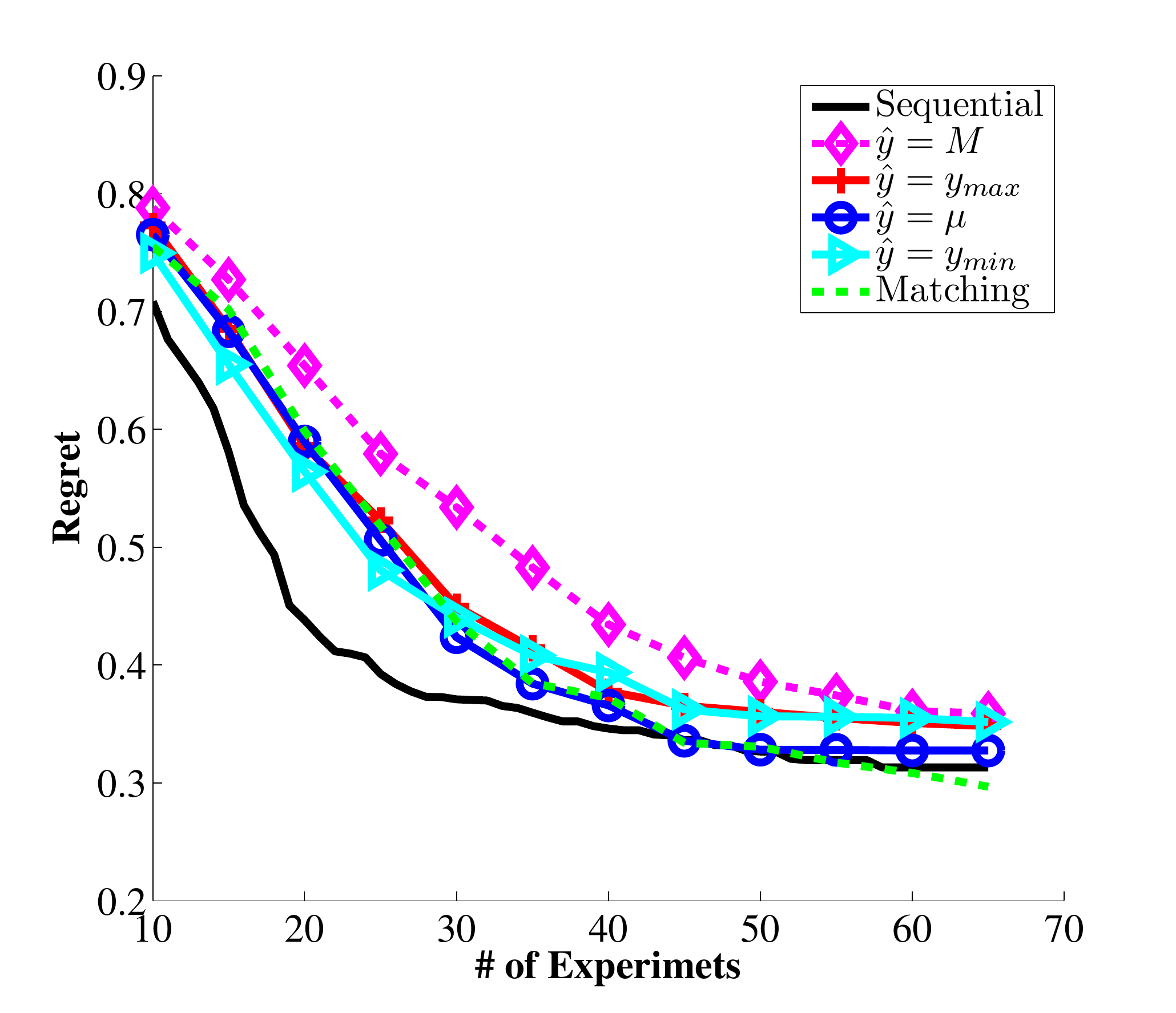} &
\includegraphics[width=1.5 in,height=1.35 in]{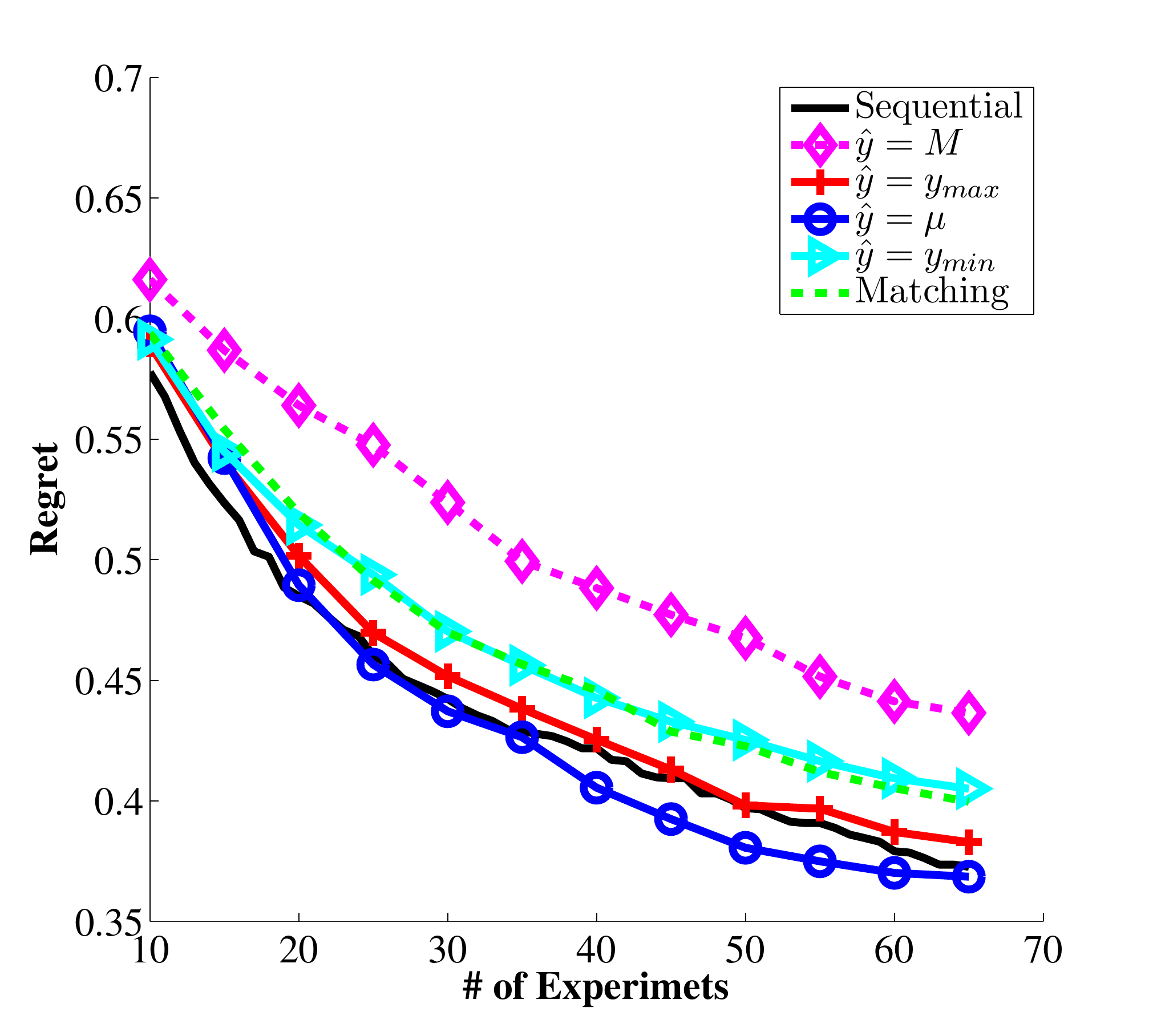} &
\includegraphics[width=1.5 in,height=1.35 in]{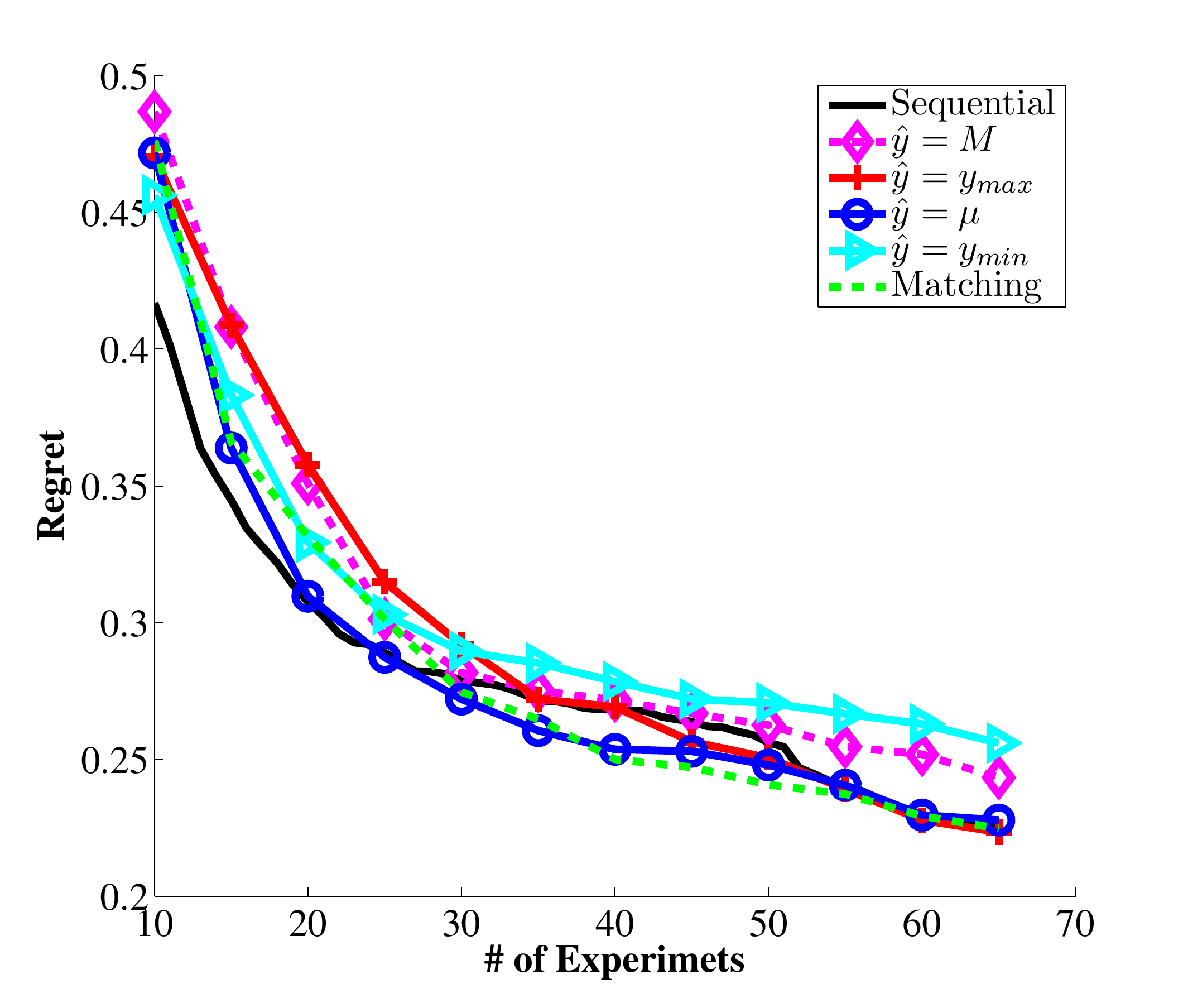} \\
Hartman(3) &Shekel &Michalewicz& Hartman(6)\\
\end{tabular}
\end{center}
\vspace{-1.5em}
\caption{The performance of different batch algorithms for batch size $5$.}
\label{fig:evaluation1}
\end{figure*}

The speedup of our proposed approach is calculated as the percentage of the samples in the whole experiment that are selected in batch mode. More specifically, if we finish $n_l$ samples in $T$ steps, the speedup is calculated as $1-\frac{T}{n_l}$. Clearly, the maximum speedup in our setting is $\% 80$, that can be only achieved if we select $5$ experiments at each time steps. For example, the speedup of proposed baseline batch approaches, Matching and CL($\hat{\mu}$), are $\% 80$. Table \ref{table:perf} shows the result.

Interestingly, all of the $6$ considered estimators achieved similar performance (comparable to EI) in terms of their regrets. The key difference between the different estimators is the level of speedup they achieve. In particular, we observe that the most speedup is achieved by $\hat{y}=\widehat{\mu}_{x|\mathcal{O}}$, for which we are able to produce over $70 \%$ speedup (very close to fully batch) for the three high dimensional functions Michalewicz, Shekel and Hartman 6. 

Further inspection of the speedup rates reveal that setting $\hat{y}$ to a large value, for example $M$, $y_{max}$, and $(1+\zeta)y_{max}$, generally leads to less speedup than the other choices. This can be explained by noting that a large value of $\hat{y}$ will lead to higher chance of violating the condition required for making the next experiment selection in Algorithm 1, which is stated in Equation 3. In particular, for a large $\hat{y}$, the next point selected by EI will most likely be very close to $x$, since the mean of the points close to $x$ are high. This will lead to a large $\gamma_z$. Further, the quantity $\|\widehat{y} - \mu_{\boldsymbol{x}|\mathcal{O}}\|_2$ is likely very large. Consequently, it is easy to violate this condition thus stop the selection process early on. In contrast, if $\hat{y}=y_{min}$, although $\|\widehat{y} - \mu_{\boldsymbol{x}|\mathcal{O}}\|_2$ is large, we expect $\gamma_z$ to be small because the next point $z$ selected by EI will likely to be far away from $x$ since the mean and variance of the points close to $x$ are very small. Considering the two terms jointly,  we expect to achieve a higher speedup by setting $\hat{y}=y_{min}$ comparing to setting $\hat{y}$ to a large value, which is exactly what we observe in our experiments. Finally, by setting $\hat{y}$ to $\mu_{\boldsymbol{x}|\mathcal{O}}$, we have $\|\widehat{y}- \mu_{\boldsymbol{x}|\mathcal{O}}\|_2=0$ and the stopping criterion only depends on $\gamma_z\theta_{\boldsymbol{x}}$. Thus we expect to achieve the maximum speedup among the different choices we consider for $\hat{y}$.

Our experimental investigation shows that the size of the batch generally increases as the experiment goes forward. This is consistent with our theoretical results in which the value of $\gamma_z\left(\theta_{\boldsymbol{x}}+\|\widehat{y} - \mu_{\boldsymbol{x}|\mathcal{O}}\|_2\right)$ decreases as the variances decreases. Note that, sampling at any arbitrary point when the number of observations is small would change the variance of the input space significantly comparing to the case where there are a lot of observation points. Therefore, the stopping criteria of Algorithm 1 is less likely to be met in the early stages of the experimental procedure where there are a few observation points.



\textbf{The $\mu$-Constant Batch Approach.}
This part of the experiments is motivated by our theoretical analysis and the goal is to shed some lights on a batch method recently proposed by \citet{Gins10}, which selects a batch of experiments that jointly maximize the EI objective. They show that finding such a batch of experiments is practically intractable. Therefore, they introduced a heuristic approach called \emph{Constant liar} to select a batch of $k$ experiments. After selecting the first experiment, \emph{Constant liar} sets the output of the selected experiment as a constant value $c$. That experiment is then added to the set of observations and the next experiment is selected. This procedure is repeated until $k$ experiments are selected. They introduced several possible ways for setting $c$, including $c=M$, $c=\widehat{\mu}$ and $c=y_{min}$. They empirically demonstrated that setting $c=M$ provided them a good result for their particular test functions. However, there is no theoretical justification or guidance toward what is the best $c$.

\begin{figure*}
\begin{center}
\begin{tabular}{@{}c@{} @{\ }c@{} @{\ }c@{} @{}c@{}}
\includegraphics[width=1.5 in,height=1.35 in]{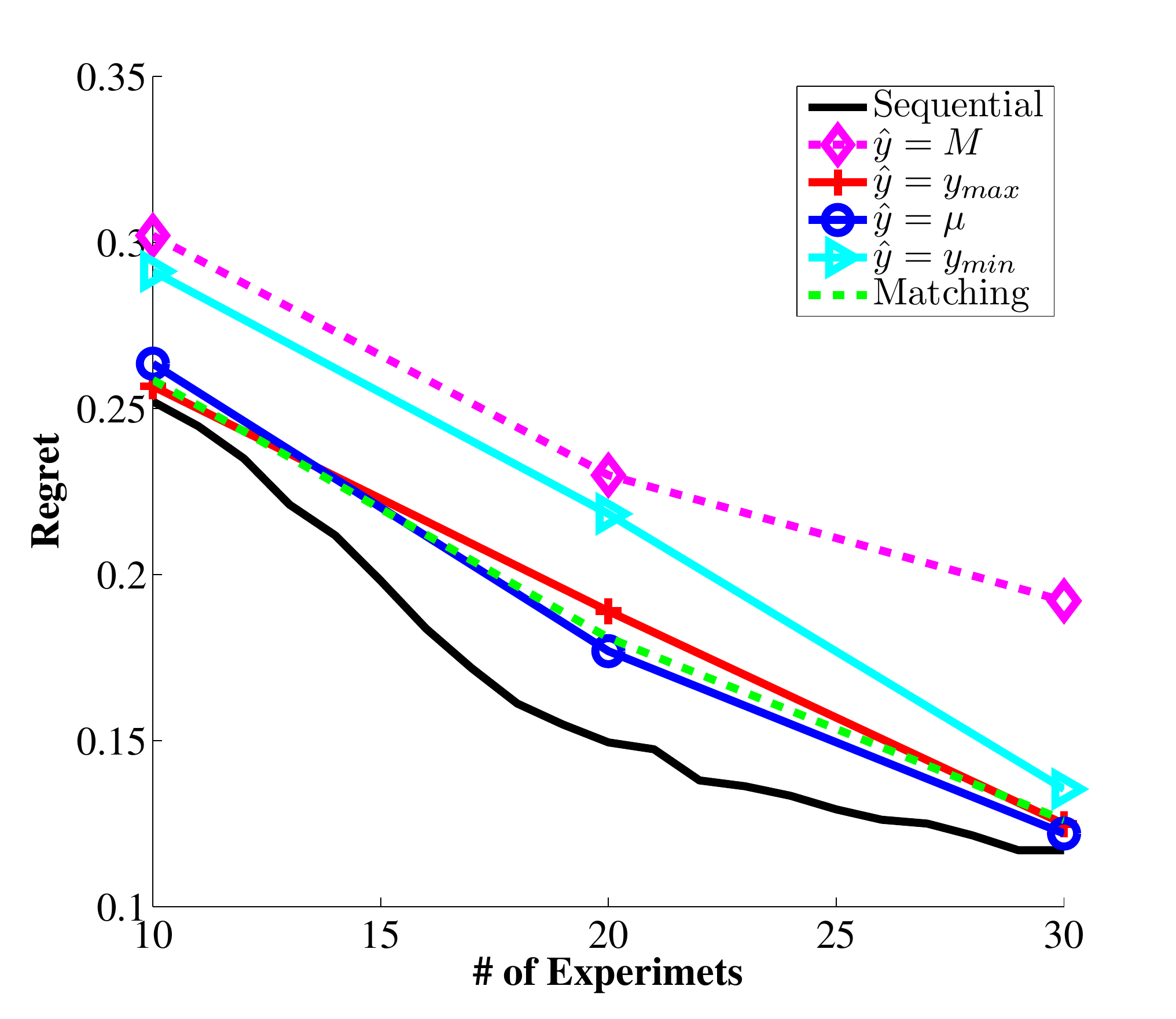} &
\includegraphics[width=1.5 in,height=1.35 in]{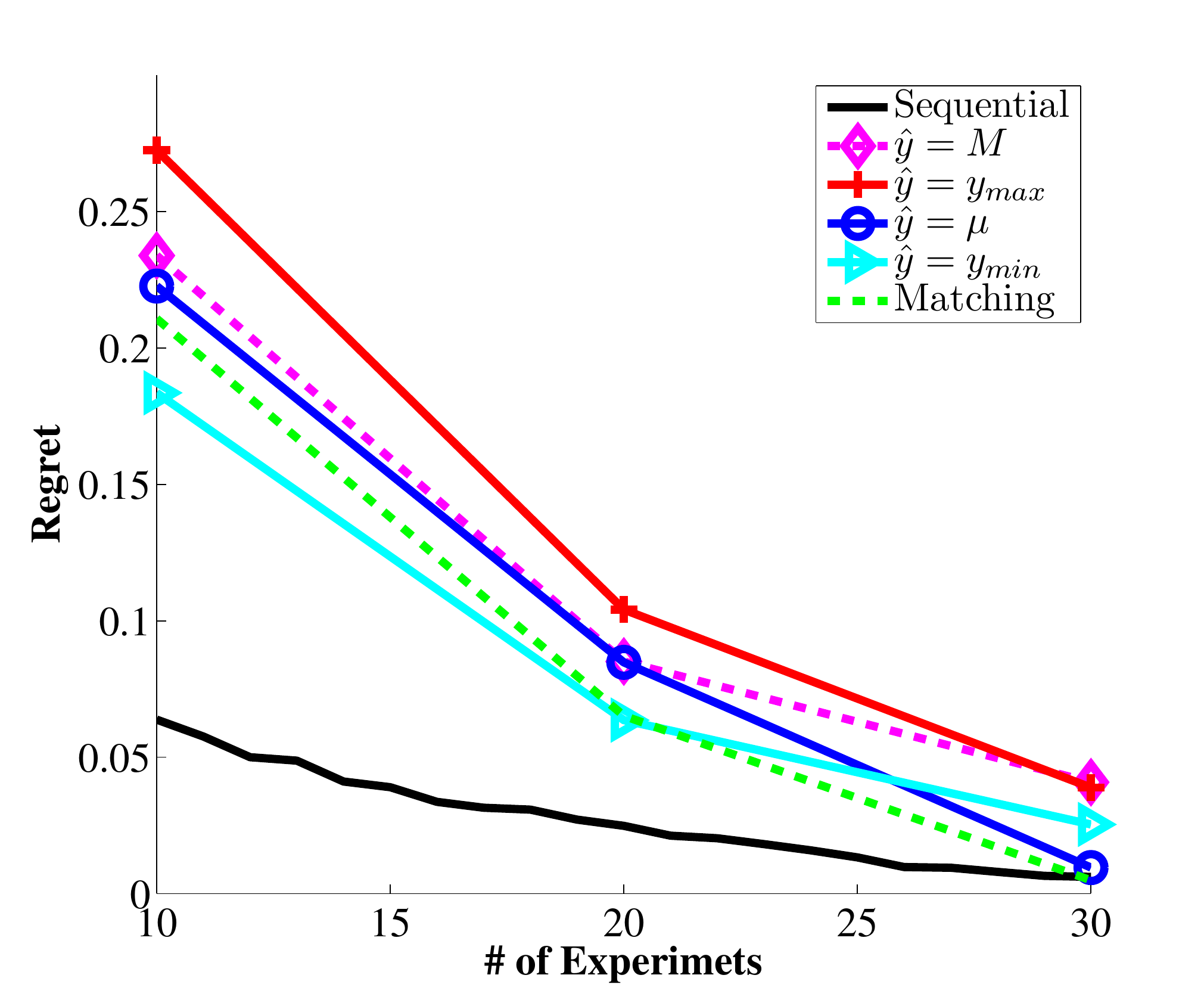} &
\includegraphics[width=1.5 in,height=1.35 in]{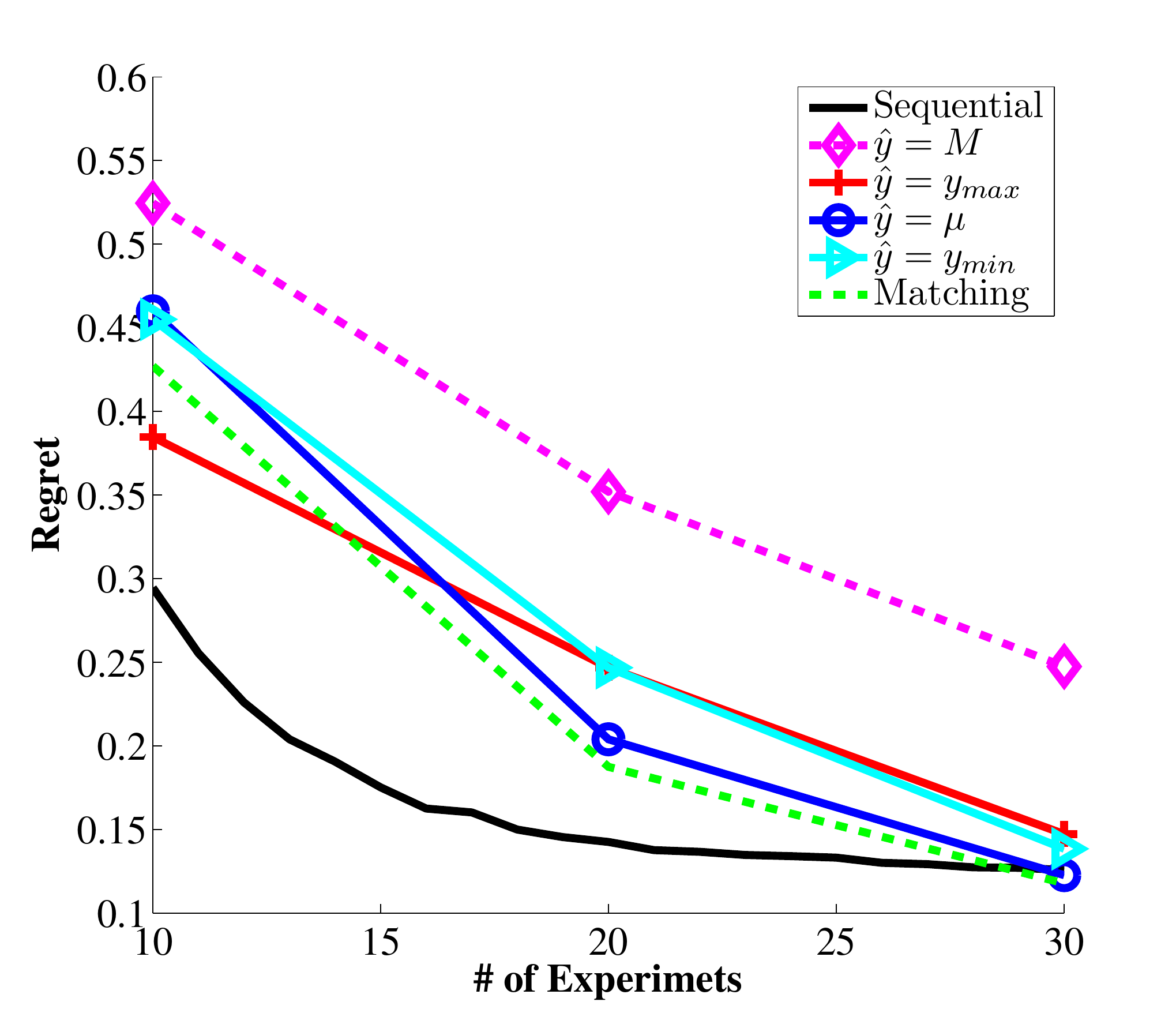} &
\includegraphics[width=1.5 in,height=1.35 in]{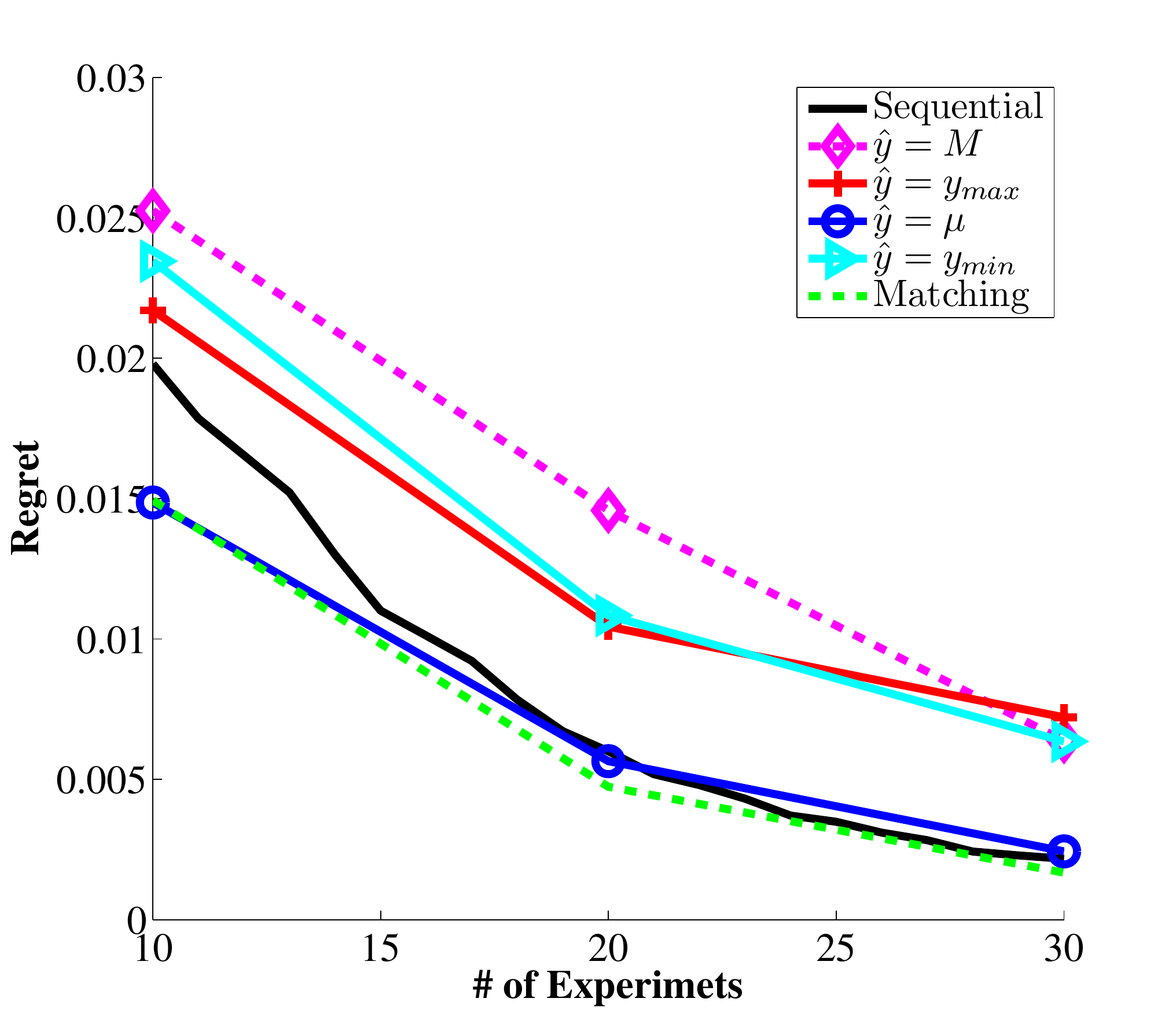} \\
Fuel Cell &Hydrogen &Cosines& Rosenbrock\\
\includegraphics[width=1.5 in,height=1.35 in]{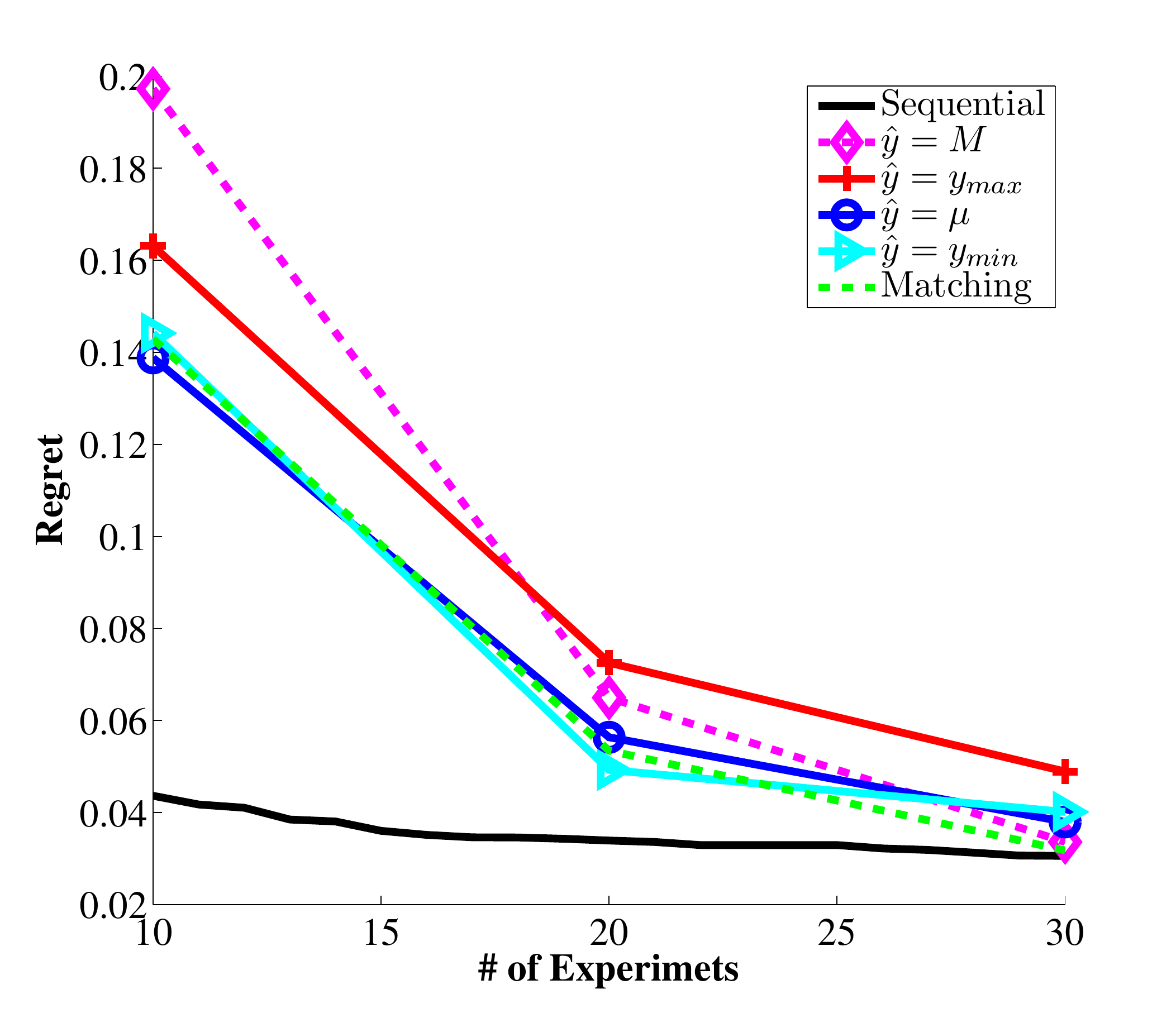} &
\includegraphics[width=1.5 in,height=1.35 in]{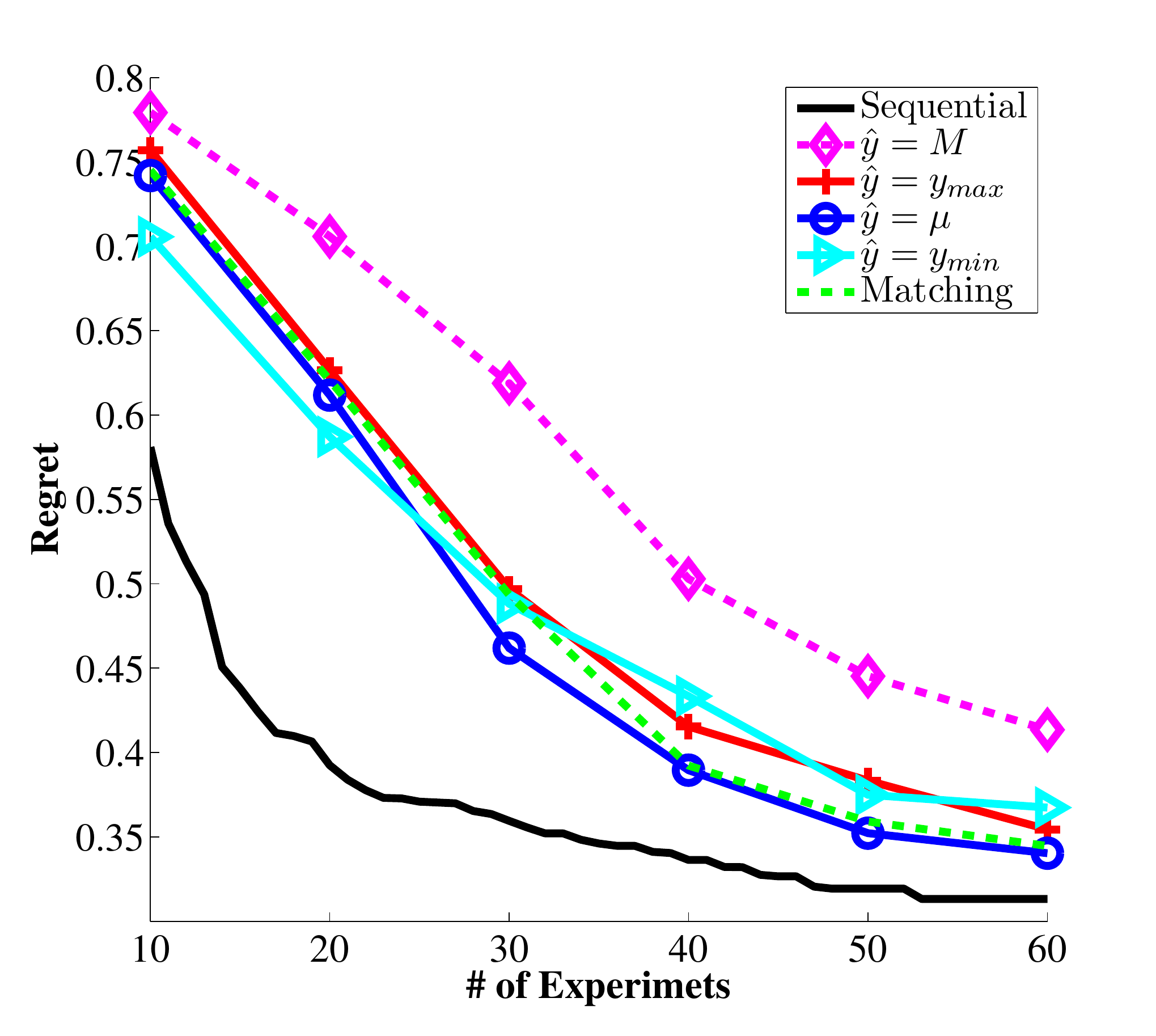} &
\includegraphics[width=1.5 in,height=1.35 in]{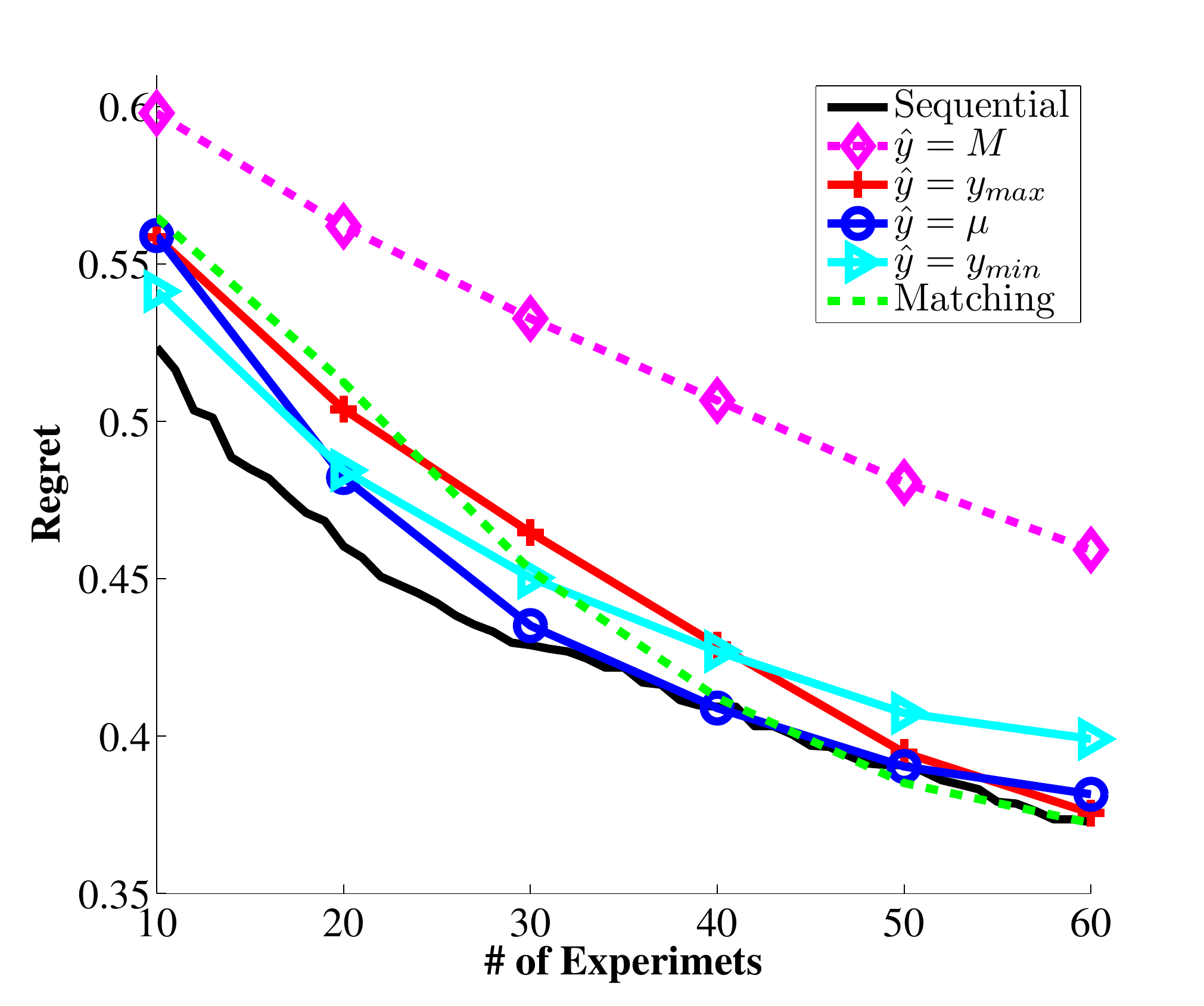} &
\includegraphics[width=1.5 in,height=1.35 in]{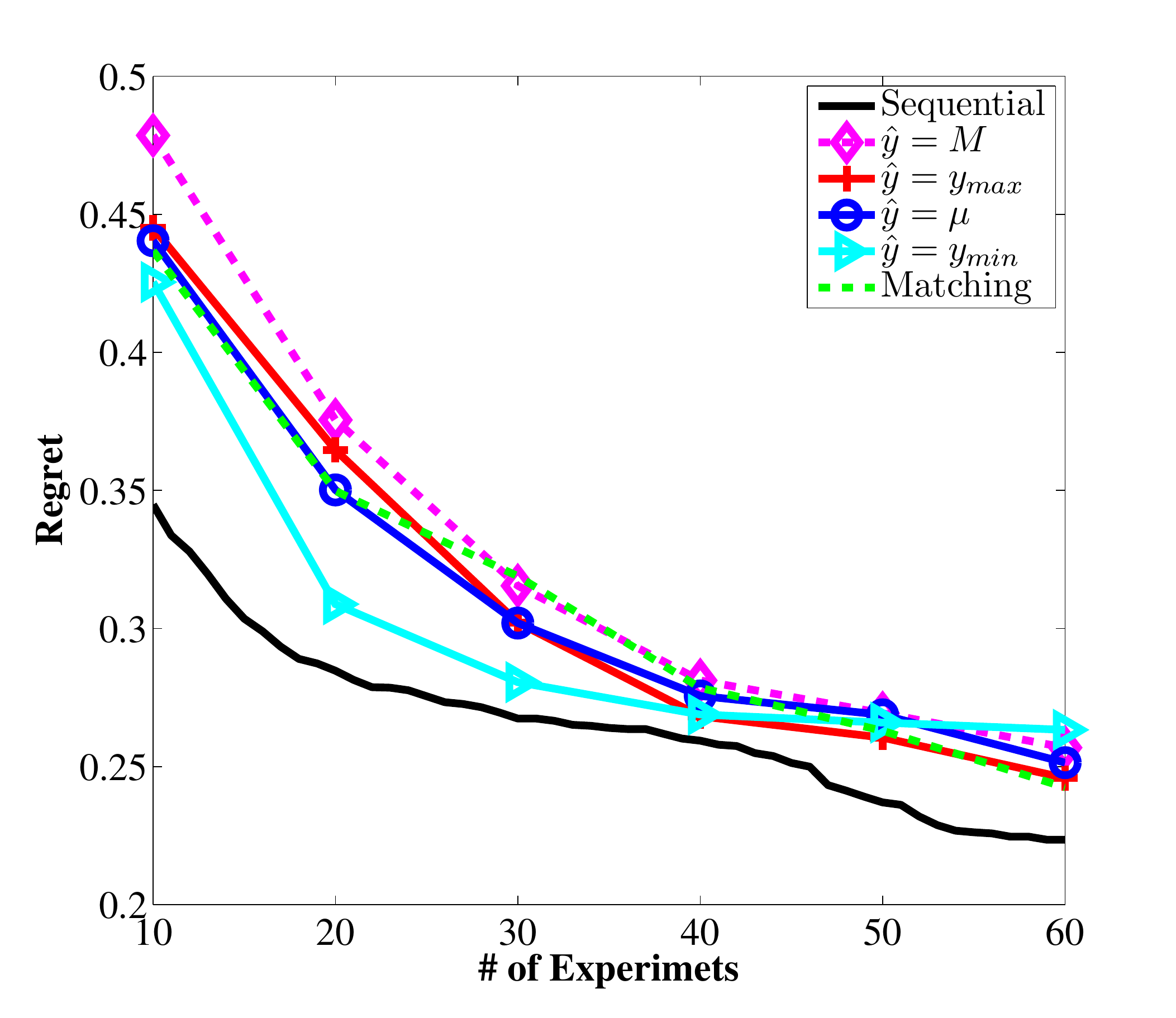} \\
Hartman(3) &Shekel &Michalewicz& Hartman(6)\\
\end{tabular}
\end{center}
\vspace{-1.5em}
\caption{The performance of different batch algorithms for batch size $10$.}
\label{fig:evaluation2}
\end{figure*}

Our theoretical analysis, in particular Corollary 1, indicates that by setting $c$ ($\hat{y}$ in this paper) to $\widehat{\mu}_{\boldsymbol{x}|\mathcal{O}}$, the condition for continued experiment selection can be easily met comparing to other settings, i.e., $\gamma_z\theta_{\boldsymbol{x}}\leq\epsilon$. Thus, a batch of $k\geq 1$ experiments are requested at most iterations without degrading the performance. This theoretical result also justifies the choice of setting $c=\widehat{\mu}_{\boldsymbol{x}|\mathcal{O}}$ in the \emph{constant liar} approach. We call this approach $\mu$-Constant Batch.  We run this algorithm on proposed $8$ benchmarks for different batch sizes $5$ and $10$. Figures \ref{fig:evaluation1} and \ref{fig:evaluation2} show the performance of $\mu$-Constant along with $5$ competitive approaches: 1) Sequential EI; 2) \emph{Constant liar} with $\hat{y}=M$; 3) \emph{Constant liar} with $\hat{y}=y_{max}$; 4) \emph{Constant liar} with $\hat{y}=y_{min}$; and 5) Matching, which is a recently proposed approach by \citet{azimi10}. For this set of experiments, we use the same experimental setup as used in Table \ref{table:perf}.

The results show that the $\mu$-constant batch approach performs very competitively compared to the Matching approach, which is one of the best existing batch Bayesian optimization approach in the literature. In addition, it is more practical than the Matching approach for high dimensional applications since its computational complexity is significantly less than the Matching algorithm. Note that the performance of $\mu$-Constant is also shown in Table \ref{table:perf} as CL($\hat{\mu}$). It is worth emphasizing that while $\mu$-Constant achieves highly competitive batch performance, it is consistently worse than sequential EI and the proposed Hybrid Batch EI algorithm. This result suggests that the stopping criterion used in Algorithm 1 is in fact effective toward identifying the condition under which we must stop increasing the batch size to avoid significant performance degradation compared to the sequential EI.


\section{Conclusion}
\label{sec:con}
In the Bayesian optimization framework, we investigated the problem of batch query selection with the goal of maintaining the performance of a sequential policy which using fewer iterations. Although our results are for general BO problems, for the sake of clarity, we focused on the task of maximizing an unknown non-convex/concave function. There are two main contributions in this paper.

Firstly, we introduce a systematic way to analyze the performance and limits of simulation-based batch BO methods by a) proving universal bounds on the bias caused by the simulation (estimation-of-outcome) error; and b) analyzing the selection of the second experiment when we have an estimate of the outcome of the first experiment. In all cases, we provide theoretical bounds on the error, relating the simulation error to the prediction error of the next best experiment.

Secondly, based on the analysis above, we proposed an algorithm that behaves optimally in expectation. This algorithm at each step decides whether or not to pick another query to add to the current batch, and as such dynamically determines the appropriate batch size at each step. In early iterations, our algorithm behaves more similar to the sequential policy and gradually moves toward a batch policy with variable batch sizes.

The empirical evaluation over both synthetic and real data shows substantial speedup (up to 78\% ) compared to the corresponding sequential policy, with little to nothing loss in the optimization performance. Our theoretical results also shed some interesting light on the \emph{Constant-liar} approach, a recently proposed batch selection method based on the EI objective.

\bibliographystyle{natbib}
\bibliography{DynamicBOarXiv}

\newpage
\appendix
\section{Proof of Theorem 1}
Recalling the notation introduced in the Theorem statement, we have
\begin{equation}
\begin{aligned}
\Delta(\sigma_z) &= CA^{-1}C^T - [C\quad k(z,\boldsymbol{x})]\left[
\begin{array}{cc}
A &  B^T \\B& k(\boldsymbol{x},\boldsymbol{x}) \end{array}\right]^{-1}\left[
\begin{array}{c}
C^T \\k(z,\boldsymbol{x}) \end{array}\right]\\
&= CA^{-1}C^T - [C\quad k(z,\boldsymbol{x})]\left[\begin{array}{cc} A^{-1}+A^{-1}B^TDBA^{-1}& -A^{-1}B^TD\\ -DBA^{-1}& D \end{array}\right]\left[
\begin{array}{c}
C^T \\k(z,\boldsymbol{x}) \end{array}\right]\\
&=\left(CA^{-1}B^T-k(z,\boldsymbol{x})\right)D\left(BA^{-1}C^T-k(z,\boldsymbol{x})\right)^T.\\ 
\end{aligned}
\nonumber
\end{equation}
This concludes the proof of the theorem.\\

\section{Proof of Theorem 2}
By definition and block matrix inversion lemma, we have
\begin{equation}
\begin{aligned}
\mu_{z|\mathcal{O},\boldsymbol{x}} - \widehat{\mu}_{z|\mathcal{O},\boldsymbol{x}} &=  k(z,\{\boldsymbol{x}_\mathcal{O},\boldsymbol{x}\}) k(\{\boldsymbol{x}_\mathcal{O},\boldsymbol{x}\},\{\boldsymbol{x}_\mathcal{O},\boldsymbol{x}\})^{-1} \left[\begin{aligned} &\quad {\bf 0}\\&\boldsymbol{y}-\widehat{\boldsymbol{y}}\end{aligned}\right]\\
&=(k(z,\boldsymbol{x}) - CA^{-1}B^T)D(\boldsymbol{y}-\widehat{\boldsymbol{y}}).
\end{aligned}
\nonumber
\end{equation}

For the second part, we have
\begin{equation}
\begin{aligned}
\mu_{z|\mathcal{O}}-\mu_{z|\mathcal{O},\boldsymbol{x}} &= CA^{-1}\boldsymbol{y}_{\mathcal{O}} - [C\quad k(z,\boldsymbol{x})]\left[
\begin{array}{cc}
A &  B^T \\B& k(\boldsymbol{x},\boldsymbol{x}) \end{array}\right]^{-1}\left[
\begin{array}{c}
\boldsymbol{y}_{\mathcal{O}} \\\boldsymbol{y} \end{array}\right]\\
&= CA^{-1}\boldsymbol{y}_{\mathcal{O}} - [C\quad k(z,\boldsymbol{x})]\left[
\begin{array}{cc}
A^{-1}+A^{-1}B^TDBA^{-1}& -A^{-1}B^TD\\ -DBA^{-1}& D \end{array}\right]\left[
\begin{array}{c}
\boldsymbol{y}_{\mathcal{O}} \\\boldsymbol{y}^{*} \end{array}\right]\\
&=\left(CA^{-1}B^T-k(z,\boldsymbol{x})\right)D\left(BA^{-1}\boldsymbol{y}_{\mathcal{O}}- \boldsymbol{y}\right)\\
&=\left(CA^{-1}B^T-k(z,\boldsymbol{x})\right)D\left(\mu_{\boldsymbol{x}|\mathcal{O}}- \boldsymbol{y}\right).\\ 
\end{aligned}
\nonumber
\end{equation}
This concludes the proof of the theorem.\\

\section{Proof of Lemma 1}
Let $\Delta_z = \max(y_{max},y_1^*)-\mu_{z|\mathcal{O},x_1^*}$. Using Theorem 2, we have
\begin{equation}
\begin{aligned}
\widehat{\Delta}_z &:= \max(y_{max},\hat{y}_1)-\widehat{\mu}_{z|\mathcal{O},x_1^*}\\
&= \max(y_{max},y_1^*) - \mu_{z|\mathcal{O},x_1^*} + \max(y_{max},\hat{y}_1) - \max(y_{max},y_1^*)\\ &\qquad\qquad\qquad\qquad - \frac{1}{\sigma_{x_1^*|\mathcal{O}}^2}\Big(k(z,x_1^*) - k(z,\boldsymbol{x}_\mathcal{O}) k(\boldsymbol{x}_\mathcal{O},\boldsymbol{x}_\mathcal{O})^{-1}k(\boldsymbol{x}_\mathcal{O},x_1^*)\Big) \big(\hat{y}_1-y_1^*\big)\\
&= \Delta_z + \underbrace{\max(y_{max},\hat{y}_1) - \max(y_{max},y_1^*) - \rho_{z,x_1^*}\frac{\sigma_{z|\mathcal{O}}}{\sigma_{x_1^*|\mathcal{O}}}\big(\hat{y}_1-y_1^*\big)}_{\delta_z}\\
&= \Delta_z + \delta_z.
\end{aligned}
\nonumber
\end{equation}
Here, $\rho_{z,x_1^*}$ represents the correlation coefficient between $x$ and $x_1$. Thus, we have $$|\delta_z|\leq\left(1+\frac{\sigma_{z|\mathcal{O}}}{\sigma_{x_1^*|\mathcal{O}}}\right)\left|\hat{y}_1-y_1^*\right|.$$

By mean-value theorem, there exists $\alpha\in[0,1]$, such that
\begin{equation}
\begin{aligned}
\underbrace{-\widehat{\Delta}_{z}\Phi\left(-\frac{\widehat{\Delta}_{z}}{\sigma_{z|\mathcal{O},x_1^*}}\right) + \sigma_{z|\mathcal{O},x_1^*} \phi\left(\frac{\widehat{\Delta}_{x}}{\sigma_{z|\mathcal{O},x_1^*}}\right)}_{\widehat{EI}(z)} &= \underbrace{-\Delta_{x}\Phi\left(-\frac{\Delta_{x}}{\sigma_{z|\mathcal{O},x_1^*}}\right) + \sigma_{z|\mathcal{O},x_1^*} \phi\left(\frac{\Delta_{x}}{\sigma_{z|\mathcal{O},x_1^*}}\right)}_{EI(z)} - \Phi\left(-\frac{\Delta_{z}+\alpha\delta_{z}}{\sigma_{z|\mathcal{O},x_1^*}}\right)\delta_{z}\\
\end{aligned}
\nonumber
\end{equation}

Thus,
\begin{equation}
\begin{aligned}
\Big|EI(z) - \widehat{EI}(z)\Big| &= \Phi\left(-\frac{\Delta_{z}+\alpha\delta_{z}}{\sigma_{z|\mathcal{O},x_1^*}}\right) \Big|\delta_{z}\Big|\\
&\leq \frac{1}{2}\Big|\delta_{z}\Big| \leq \frac{1}{2}\left(1+\frac{\sigma_{z|\mathcal{O}}}{\sigma_{x_1^*|\mathcal{O}}}\right)\Big|\hat{y}_1-y_1^*\Big|.
\end{aligned}
\nonumber
\end{equation}

This concludes the Proof of Lemma.\\

\section{Proof of Theorem 3}
By optimality of $x_2$ and $x_2^*$, we have
\begin{equation}
EI(x_2) - \widehat{EI}(x_2) \leq EI(x_2^*) - \widehat{EI}(x_2) \leq EI(x_2^*) - \widehat{EI}(x_2^*).\\
\nonumber
\end{equation}

Using Lemma 1, we get
\begin{equation}
\begin{aligned}
\Big| EI(x_2^*) - \widehat{EI}(x_2)\Big| &\leq\frac{1}{2}\left(1+\frac{\max(\sigma_{x_2|\mathcal{O}},\sigma_{x_2^*|\mathcal{O}})} {\sigma_{x_1^*|\mathcal{O}}}\right)\Big|\hat{y}_1-y_1^*\Big|.
\end{aligned}
\nonumber
\end{equation}

We can continue
\begin{equation}
\begin{aligned}
\widehat{EI}(x_2) - \widehat{EI}(x_2^*) & \leq  \Big| \widehat{EI}(x_2) - EI(x_2^*)\Big| + \Big| EI(x_2^*) - \widehat{EI}(x_2^*)\Big|\\
&\leq \left(1+\frac{\max(\sigma_{x_2|\mathcal{O}},\sigma_{x_2^*|\mathcal{O}})} {\sigma_{x_1^*|\mathcal{O}}}\right)\Big|\hat{y}_1-y_1^*\Big|
\end{aligned}
\nonumber
\end{equation}

By optimality of $x_2^*$, the derivative of EI is zero at $x_2^*$ and Taylor series expansion yields that for some $\alpha\in [0,1]$, we have
\begin{equation}
\begin{aligned}
\widehat{EI}(x_2^*) - \widehat{EI}(x_2) &=  \frac{1}{2} (x_2^*-x_2)^T\;\frac{d^2\widehat{EI}}{dx^2}\Big((1-\alpha)x_2^*+\alpha x_2\Big)\;(x_2^*-x_2).
\end{aligned}
\nonumber
\end{equation}

Finally, we get
\begin{equation}
\begin{aligned}
\Big\|x_2^* - x_2\Big\|_2^2 &\leq \frac{2\Big|\widehat{EI}(x_2^*) - \widehat{EI}(x_2)\Big|}{\Sigma_{\min}\!\!\left(\frac{d^2\widehat{EI}}{dx^2}((1-\alpha)x_2^*+\alpha x_2)\right)}\\ &\leq \frac{2}{\Sigma_{\min}}\left(1+\frac{\max(\sigma_{x_2|\mathcal{O}},\sigma_{x_2^*|\mathcal{O}})} {\sigma_{x_1^*|\mathcal{O}}}\right)\Big|\hat{y}_1-y_1^*\Big|.
\end{aligned}
\end{equation}

\section{Proof of Corollary 2}
From theorem \ref{theorem:varchange}, there is an interesting finding which shows that the difference of variance of any point $z$ in the input space after adding the point $x^*$ to our observation set is exactly $D\left(k(z,x_1^*)-BA^{-1}C^T\right)^2$ if we consider $x^*_1$ as a single point. Since $\delta_z^2-\delta^{2*}_z>0$, therefore $m\geq 0$.  In addition, when $|x^*|=1$, it can be shown that $m^{-1}=\sigma^{*2}$.
Thus, we are interested in the points where $\delta_z^2-\delta^{*2}\geq\epsilon \geq 0$. Therefore we have:
\begin{equation}
\begin{aligned}
&\delta_z^2-\delta^{2*}_z -\epsilon\geq 0 \\
&Dk(x_1^*,z)^2-\left(2DCA^{-1}B^T\right)k(x_1^*,z)+\left(D(CA^{-1}B^T)^2\right)-\epsilon\geq 0\\
\end{aligned}
\end{equation}
this is a quadratic function  of $k(x_1^*,z)$ with 2 real roots as follow:
\begin{equation}
\begin{aligned}
k(x_1^*,z)=\left\{\begin{array}{r} r_1=CA^{-1}B^T+\sqrt{\frac{\epsilon}{D}}\\
r_2=CA^{-1}B^T-\sqrt{\frac{\epsilon}{D}}
\end{array}\right.
\end{aligned}
\end{equation}

So we are interested in the region where $k(x_1^*,z)\geq r_1$ or $k(x_1^*,z) \leq r_2$. For large value of $\epsilon$ the $r_2<0$ and since $k(x_1^*,z)>0$,  we are only interested in where $k(x_1^*,z)\geq r_1$. Therefore we have
\begin{equation}
\begin{aligned}
1 \geq k(x_1^*,z)=e^{\frac{-\parallel z-x^*_1 \parallel^2}{l}}&\geq CA^{-1}B^T+\sqrt{\frac{\epsilon}{D}}\geq 0\\
\end{aligned}
\end{equation} 

We are trying to introduce an upper bound for $r_1$ which is free from $P_z$. Clearly $CA^{-1}B^T\leq |CA^{-1}B^T|$. Then we have, 
\begin{equation}
\begin{aligned}
|CA^{-1}B^T|&=\parallel CA^{-1}B^T\parallel_2\\
&\leq \parallel C\parallel_2\;\parallel A^{-1}B^T\parallel_2 \qquad  \textbf{Cauchy-Shwrz inequality}\\
&\leq \sqrt{n} \parallel C\parallel_{\infty}\;\parallel A^{-1}B^T\parallel_2\\
&\leq \sqrt{n} \parallel A^{-1}B^T\parallel_2 \qquad  \textbf{sinec $0\leq \parallel C\parallel_{\infty}\leq 1$ }\\
\end{aligned}
\end{equation} 
Therefore we are certain about the point satisfying the following equation
\begin{equation}
\begin{aligned}
k(x_1^*,z) &\geq \sqrt{n}\parallel A^{-1}B^T\parallel_{2} +\sqrt{\frac{\epsilon}{D}}\\
\parallel z-x^* \parallel^2&\leq -l \ln \left(\sqrt{n} \parallel A^{-1}B^T\parallel_{2} +\sqrt{\frac{\epsilon}{D}}\right)\\
\parallel z-x^* \parallel^2&\leq -l \ln \left( \sqrt{n}\parallel A^{-1}B^T\parallel_{2} +\sigma^*\sqrt{\epsilon}\right)\\
\end{aligned}
\end{equation}

\section{Proof of Corollary 3}
\begin{equation}
\begin{aligned}
\left\|\left(CA^{-1}B^T-k(x_1^*,z)\right)D\right\|_\infty  \sqrt{\frac{2}{\pi}}\|\boldsymbol{\sigma_{x_1^*|\mathcal{O}}}\|_1&\geq \epsilon\\
 \left |\left(CA^{-1}B^T-k(x_1^*,z)\right)\right | & \geq \frac{\epsilon}{\sqrt{\frac{2}{\pi}} |\boldsymbol{\sigma_{x_1^*|\mathcal{O}}}|D} \\
|(CA^{-1}B^T-k(x_1^*,z))|^2&\geq \left(\frac{\epsilon}{\sqrt{\frac{2}{\pi}} |\boldsymbol{\sigma_{x_1^*|\mathcal{O}}}|D}\right)^2\\
(CA^{-1}B^T)^2+k(x_1^*,z)^2 &\geq \frac{\pi\epsilon^2}{2\sigma_{x_1^*|\mathcal{O}}^2D^2}\\
k(x_1^*,z)^2 &\geq \frac{\pi\epsilon^2}{2\sigma^6_{x_1^*|\mathcal{O}}}-n\parallel A^{-1}B^T\parallel_{2}^2\\
\parallel z-x^*\parallel^2&\leq -l\ln \sqrt{\frac{\pi\epsilon^2}{2\sigma^6_{x_1^*|\mathcal{O}}}-n\parallel A^{-1}B^T\parallel_{2}^2} 
\end{aligned}
\end{equation}
Note that  $|a-b|^2\leq 2*(a^2+b^2)$.  Therefore $\mathbb{E}[|\mu_{z|\mathcal{O},\boldsymbol{x}} - \widehat{\mu}_{z|\mathcal{O},\boldsymbol{x}}|] \geq\epsilon$ if we have 

\begin{equation}
\parallel z-x^*\parallel^2\leq -l\ln \sqrt{\frac{\pi\epsilon^2}{2\sigma^6_{x_1^*|\mathcal{O}}}-n\parallel A^{-1}B^T\parallel_{2}^2}  
\end{equation}

\end{document}